\documentclass[lettersize,journal]{IEEEtran}
\ifCLASSOPTIONcompsoc
  \usepackage[nocompress]{cite}
\else
  \usepackage{cite}
\fi
\ifCLASSINFOpdf
\else
\fi
\usepackage{bbding}

\usepackage{amsfonts}

\usepackage{amsmath}

\usepackage{graphicx}

\usepackage{float}  

\usepackage{subfigure}

\usepackage{bm}

\usepackage{courier}

\usepackage{booktabs}

\usepackage{multirow}

\usepackage{rotating}

\usepackage{amsthm}

\usepackage{footnote}

\usepackage{color}

\usepackage{stfloats}

\usepackage{threeparttable}

\usepackage{adjustbox}

\usepackage{amssymb}

\usepackage[table,dvipsnames]{xcolor}

\newtheorem{definition}{Definition}

\newtheorem{property}{Property}

\newtheorem{conjecture}{Conjecture}

\newtheorem{main result}{Main Result}

\newtheorem{formulation}{Formulation}

\usepackage{color}

\usepackage[skins]{tcolorbox}
\tcbuselibrary{breakable}

\definecolor{mygray}{gray}{.85}
\definecolor{mygray1}{gray}{.7}
\definecolor{mygray2}{gray}{.93}

\makeatletter
\newcommand{\thickhline}{%
	\noalign {\ifnum 0=`}\fi \hrule height 1pt
	\futurelet \reserved@a \@xhline
}
\makeatother

\begin{document}
%
\title{Spatial-Frequency Discriminability for Revealing Adversarial Perturbations}
%
%
%
%

\author{Chao~Wang,~Shuren~Qi,~Zhiqiu~Huang,~Yushu~Zhang,~\IEEEmembership{Senior~Member,~IEEE},\\~Rushi~Lan,~Xiaochun~Cao,~\IEEEmembership{Senior~Member,~IEEE},~and~Feng-Lei Fan~\IEEEmembership{Senior~Member,~IEEE}%

\thanks{This work was supported in part by the Joint Funds of the National Natural Science Foundation of China under Grant U2241216, and in part by the Youth Talents Support Project of Guangzhou Association for Science and Technology. \emph{(Corresponding authors: S. Qi and Z. Huang}.)}

\IEEEcompsocitemizethanks{
	\IEEEcompsocthanksitem C. Wang, S. Qi, Z. Huang, and Y. Zhang are with the College of Computer Science and Technology, Nanjing University of Aeronautics and Astronautics, Nanjing 211106, China; S. Qi is also with the Department of Mathematics, The Chinese University of Hong Kong, Hong Kong (e-mail: {c.wang, shurenqi, zqhuang, yushu}@nuaa.edu.cn).
	\IEEEcompsocthanksitem R. Lan is with the Guangxi Key Laboratory of Image and Graphic Intelligent Processing, Guilin University of Electronic Technology, Guilin 541004, China (e-mail: rslan@guet.edu.cn).
	\IEEEcompsocthanksitem X. Cao is with the School of Cyber Science and Technology, Shenzhen Campus of Sun Yat-sen University, Shenzhen 518107, China (e-mail: caoxiaochun@mail.sysu.edu.cn).
	\IEEEcompsocthanksitem F. Fan is with the Department of Mathematics, The Chinese University of Hong Kong, Hong Kong (e-mail: flfan@math.cuhk.edu.hk).}
	
\thanks{Copyright © 2024 IEEE. Personal use of this material is permitted. However, permission to use this material for any other purposes must be obtained from the IEEE by sending an email to pubs-permissions@ieee.org.}
}

%
%

\markboth{C. Wang \MakeLowercase{\textit{et al.}}: Spatial-Frequency Discriminability for Revealing Adversarial Perturbations} {C. Wang \MakeLowercase{\textit{et al.}}: Spatial-Frequency Discriminability for Revealing Adversarial Perturbations}
%



\IEEEtitleabstractindextext{%
\begin{abstract}
The vulnerability of deep neural networks to adversarial perturbations has been widely perceived in the computer vision community. From a security perspective, it poses a critical risk for modern vision systems, \emph{e.g.}, the popular Deep Learning as a Service (DLaaS) frameworks. For protecting deep models while not modifying them, current algorithms typically detect adversarial patterns through discriminative decomposition for natural and adversarial data. However, these decompositions are either biased towards frequency resolution or spatial resolution, thus failing to capture adversarial patterns comprehensively. Also, when the detector relies on few fixed features, it is practical for an adversary to fool the model while evading the detector (\emph{i.e.}, defense-aware attack). Motivated by such facts, we propose a discriminative detector relying on a spatial-frequency Krawtchouk decomposition. It expands the above works from two aspects: 1) the introduced Krawtchouk basis provides better spatial-frequency discriminability, capturing the differences between natural and adversarial data comprehensively in both spatial and frequency distributions, \emph{w.r.t.} the common trigonometric or wavelet basis; 2) the extensive features formed by the Krawtchouk decomposition allows for adaptive feature selection and secrecy mechanism, significantly increasing the difficulty of the defense-aware attack, \emph{w.r.t.} the detector with few fixed features. Theoretical and numerical analyses demonstrate the uniqueness and usefulness of our detector, exhibiting competitive scores on several deep models and image sets against a variety of adversarial attacks.
\end{abstract}

}

\maketitle

\IEEEdisplaynontitleabstractindextext

%
\IEEEpeerreviewmaketitle

\section{Introduction}\label{intro}

%
%
%
%

\IEEEPARstart{M}{odern} deep neural networks (DNN) are highly discriminative representation techniques, exhibiting impressive  performance in extensive scenarios ranging from perceptual information understanding to hard scientific problem deciphering \cite{ref1}. Besides their effectiveness, the emerging Deep Learning as a Service (DLaaS) paradigm allows developers to apply or deploy proven DNN models in a simple and efficient manner \cite{ref2}. The above two factors have led to the widespread emergence of deep learning-based artificial intelligence systems in everyday life, even expanding to many security and trust sensitive scenarios \cite{ref3}.
	
On the other hand, the robustness of DNN models has raised general concerns, especially in the computer vision community \cite{ref4}. It has been shown that adversarial perturbations on the input example can cause significant fluctuations on such deep representations, even though such perturbations are almost imperceptible for humans \cite{ref5}.

Since the first work by Szegedy \emph{et al.} \cite{ref6}, various attack methods have been proposed for well crafting such adversarial perturbations. In general, The goals of attackers cover high fooling rate \cite{ref7}, low perceptual loss \cite{ref8}, efficient generation \cite{ref9}, high transferability \emph{w.r.t.} models \cite{ref10}, high universality \emph{w.r.t.} examples \cite{ref11}, less need for model knowledge \cite{ref12}, easy physical implementation \cite{ref13}, etc. Their applications have expanded from classical classification tasks to semantic segmentation \cite{ref79}, object tracking \cite{ref80}, low-level processing \cite{ref81}, security \cite{ref82}, forensic \cite{ref83}, privacy \cite{ref84} and so on. Hence, these recent advances allow an adversary to perform effective evasion at a low cost, which interferes the deep representation and thus fundamentally destabilizes the system. In addition, the popularity of DLaaS-based development further increases this security threat. Specifically, a successful attack on few off-the-shelf deep models in DLaaS platform will affect a wide range of hosts and their systems, especially where safety-critical scenarios may be involved.

Integrating the above facts, it is generally agreed that adversarial perturbations have become a real threat that the security community has to face.

\begin{figure*}[!t]
	\centering
	\includegraphics[scale=0.6]{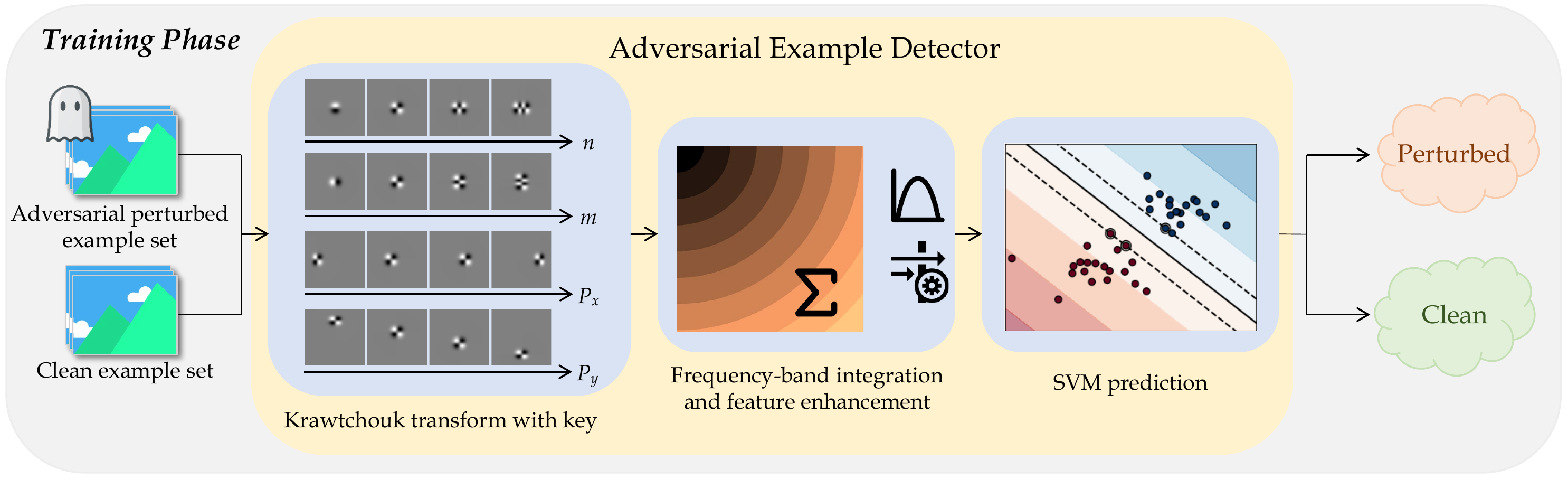}
	\centering
	\caption{Illustration for the training phase of the proposed adversarial example detector. The detector is trained on a set of adversarial/clean image examples along with corresponding labels. The detector consists of three main steps: 1) the image is projected into a space defined by Krawtchouk polynomials, where the frequency parameters $(n,m)$ and spatial parameters $({P_x},{P_y})$ are determined by key; 2) the obtained coefficients are integrated and enhanced to form a compact-but-expressive feature vector by certain beneficial priors; 3) such features are fed into an SVM for the prediction, which is the only learning part in the detector.}
\end{figure*}

\begin{figure*}[!t]
	\centering
	\includegraphics[scale=0.6]{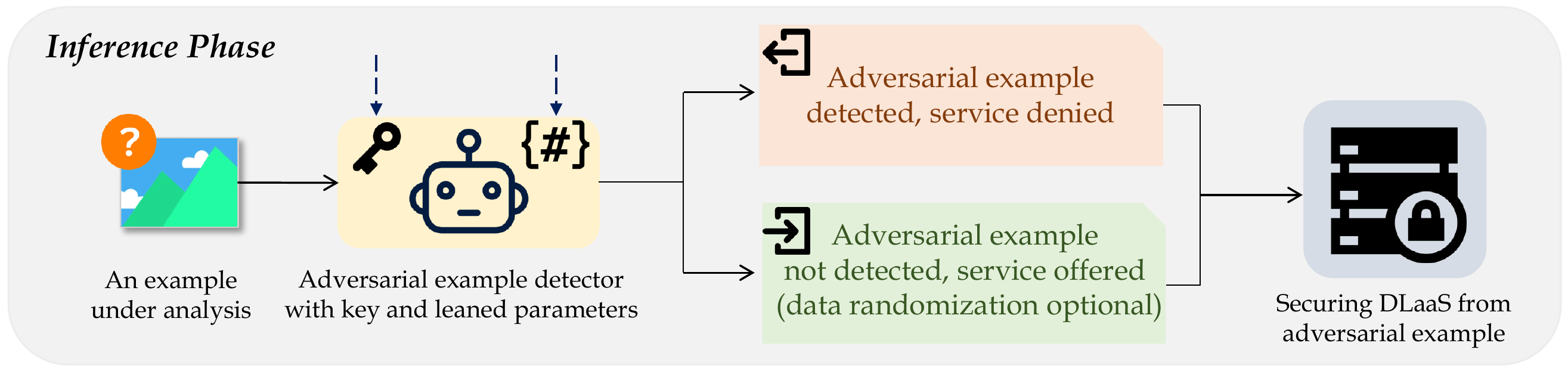}
	\centering
	\caption{Illustration for the inference phase of the proposed adversarial example detector. When the detector is trained (\emph{i.e.}, with the key and learned SVM parameters), it can be deployed in various real-world scenarios, where a DLaaS scenario is chosen as an example. For an image under analysis, our detector predicts whether it contains adversarial perturbations. With such prediction, the DLaaS is able to deny the service when the adversarial perturbation is revealed.}
\end{figure*}

\subsection{State of the Arts}\label{intro-sota}

For protecting deep neural networks and their application systems, a plethora of defense strategies towards adversarial attacks are designed over diverse research hypotheses and implementation paths.

The straightforward strategy is to include adversarial examples into the training, called \emph{adversarial training}, which promotes the model learning and adapting to adversarial patterns \cite{ref14}. Such data-level defenses are intuitively and empirically effective, but the retraining increases the implementation cost, and the resulting robustness cannot be adaptively/interpretably generalized to unseen adversarial patterns \cite{ref15}.

Another popular strategy seeks architecture-level designs as defenses, \emph{e.g.}, \emph{regularization structure} \cite{ref16, ref17, ref18} and \emph{certified robustness} \cite{ref72, ref73, ref74}, embedding adversarial robustness priors into networks. Ideally, such built-in designs have the potential to achieve adaptive/interpretable robustness against adversarial examples. Yet, in practice, tricky trade-offs between clean and robust accuracy are almost inevitable \cite{ref19}, and the additional cost for resetting DLaaS is also significant.

In contrast to the above direct attempts on robustness enhancement, the detection-only approach \cite{ref20} serve as a pre-processing for deep models to filter out potential adversarial examples, without modifying the model itself. Such an \emph{adversarial perturbation detector} can reduce both clean-accuracy loss and additional implementation costs \emph{w.r.t.} adversarial training and architecture-level design, being particularly suitable for DLaaS scenarios. Technically, such forensic detectors strongly relies on proper representation of the adversarial patterns while not being disturbed by the image contents, \emph{i.e.}, the discriminative decomposition of natural-artificial data \cite{ref21}, \cite{ref22}.

Here, the priori knowledge of adversarial perturbation is potentially useful to achieve such discriminative decomposition. As a classical prior, certain visual loss regularization forces the learning perturbation to appear mainly in high frequency bands where the human visual system is less sensitive. From this assumption, researchers have tried to extract high-frequency clues with few fixed features and non-orthogonal decomposition, such as denoising filters \cite{ref23} and Spatial Rich Model (SRM) \cite{ref24}. However, such detectors have been found to be no longer accurate and secure due to the emergence of low-frequency adversarial perturbations \cite{ref78}. Therefore, a series of orthogonal decompositions are introduced, such as Principal Component Analysis (PCA) \cite{ref25}, Discrete Cosine Transform (DCT) \cite{ref26}, Discrete Sine Transform (DST) \cite{ref27}, and Discrete Wavelet Transform (DWT) \cite{ref27}, capturing distribution differences in a wider sense than just high frequencies. (See Appendix A for a comprehensive review of related works.)

\subsection{Motivations}\label{intro-moti}

This line of adversarial perturbation detectors still falls short in accuracy and security, especially for real-world scenarios.

\begin{itemize}
	\item For the \emph{accuracy}, they are able to provide good detection performance when the scale of either the image patterns or the perturbation patterns is limited (\emph{e.g.}, MNIST with one attack), while the accuracy degrades significantly when both are large \cite{ref28}. This phenomenon implies that existing representation methods fail to capture adversarial patterns comprehensively.
\end{itemize}

\begin{itemize}
	\item As for the \emph{security}, the successful defense-aware attack that evades the detector and fools the model, is realistic when the detector relies on few fixed features. Technically, new perturbation can be learned by penalizing the specific features that the detector focuses on.
\end{itemize}

\subsection{Contributions}\label{intro-cont}

Motivated by above facts, we attempt to present a more accurate and secure adversarial example detector, technically enabled by a spatial-frequency discriminative decomposition, as shown in Figs. 1 and 2. To the best of our knowledge, this is a very early work on improving both accuracy and security of detector from the fundamental decomposition stage.

\begin{itemize}
	\item Regarding the \emph{accuracy}, we attribute the failure at larger scales to the contradiction of spatial and frequency discriminability in the decomposition, surprisingly revisiting a classical problem of signal processing \cite{ref30}. Recently, Agarwal \emph{et al.} \cite{ref27, ref28} preliminarily explored this fundamental problem, where a decision-level fusion of the DST (biased towards frequency) and DWT (biased towards spatial) was designed to mitigate such contradiction. In this paper, we introduce Krawtchouk polynomials as basis functions for the discriminative decomposition, providing a mid-scale representation different from the global trigonometric basis in DST and the local wavelet basis in DWT. Note that such a representation with rich spatial-frequency information can provide more clues of adversarial patterns for the prediction, being a more flexible detector than the decision-level fusion \cite{ref27}, \cite{ref28}.
\end{itemize}

\begin{itemize}
	\item Regarding the \emph{security}, we attribute the defense-aware attack to the accessibility of such few fixed features for the adversary, as a foundational threat in some existing methods. In this paper, we propose a confusion strategy for the adversary based on random feature selection \cite{ref31}. More specifically, a pseudorandom number generator determines the decomposition parameters, and the host controls such generator by setting the secret keys (\emph{i.e.}, the seed values). After that, the defense-aware attack becomes more difficult as the confusion of the boundary between to-be-attacked and to-be-evaded features.
\end{itemize}

\section{General Formulation}\label{gf}

In this section, we formulate the basic aspects involved in this paper, \emph{i.e.}, model, attack, and defense.

\subsection{Model Formulation}\label{gf-mod}

We focus on deep convolutional neural network models for image classification tasks. Such classification models can be formulated as a mapping ${\cal M}:{\cal X} \to {\cal Y}$, where ${\cal X} \subset {[0,1]^{W \times H \times C}}$ is the image space with the image size $W \times H \times C$ and normalized pixel intensity $[0,1]$; ${\cal Y} = \{ 1,2,...,N\} $ is the label space with the category size $N$. For a clean data point $({\bf{x}},{\bf{y}}) \in {\cal X} \times {\cal Y}$, the classification model ${\cal M}$ is considered to be correct if and only if ${\cal M}({\bf{x}}) = {\bf{y}}$.

\subsection{Attack Formulation}\label{gf-att}

\emph{Attack objective.} We focus on evasion attacks against the above image classification models. For an image ${\bf{x}}$ with the true label ${\bf{y}}$, the goal of the attacker is to find an adversarial perturbation ${\bf{\delta }}$ such that
	\begin{equation}
		{\cal M}({\bf{x}} + {\bf{\delta }}) \ne {\bf{y}},
	\end{equation}
\emph{i.e.}, fooling the prediction, typically under a norm-based constraint for the imperceptibility of perturbation:
	\begin{equation}
		||{\bf{\delta }}|| < \varepsilon.
    \end{equation}

The resulting perturbed input for the model, \emph{i.e.}, the adversarial example, is denoted as ${\bf{x'}} = {\bf{x}} + {\bf{\delta }} \in {\cal X}$. Note that the above attack objective is a formulation for the defense-unaware scenario, in line with the threat assumption of most related works. As for the defense-aware attack, the goal of the attacker include the fooling of the adversarial perturbation detector: ${\cal D}({\bf{x'}}) = {\cal D}({\bf{x}})$, in addition to the misclassiﬁcation and imperceptibility above. Here, we denote the above defense-unaware and defense-aware attacks as ${\cal A}({\bf{x}}) = {\bf{\delta }}$. More detailed formulation on attacker and defender will be provided later.

\emph{Knowledge and capability.} For the defense-unaware scenario, the attacker has perfect knowledge of the image classification model ${\cal M}$ (\emph{i.e.}, full access to its mechanism and parameters), but has zero knowledge of the detector ${\cal D}$ (or not aware of its presence). For the defense-aware scenario, the attacker likewise has perfect knowledge of the model ${\cal M}$ and has limited knowledge of the detector ${\cal D}$ — the attacker is aware that the given model ${\cal M}$ is being secured with a detector. For both scenarios, the attacker has the capability to arbitrarily modify pixels within a given image ${\bf{x}}$.

\subsection{Defense Formulation}\label{gf-def}

\emph{Defense objective.} The goal of our defense is to design an adversarial perturbation detector ${\cal D}$ such that ${\cal D}({\bf{x}}) = 0$ (\emph{i.e.}, predicted as clean example) and ${\cal D}({\bf{x'}}) = 1$ (\emph{i.e.}, predicted as adversarial example) for any clean image ${\bf{x}} \in {\cal X}$ and corresponding adversarial example ${\bf{x'}}$ by any pertinent ${\cal A}$. In other words, since ${\bf{x}}$ and ${\bf{x'}}$ differ only in ${\bf{\delta}}$, the above binary classification task is practically equivalent to a hypothesis testing for the presence of adversarial patterns (\emph{w.r.t.} ${\bf{\delta}}$) under the strong interference from image content (\emph{w.r.t.} ${\bf{x}}$).

\section{Towards Accurate and Secure Detector: Formulation}\label{main-f}

In this section, we provide new formal analyses of the concerned accuracy and security issues, as the theoretical basis for the proposed detector.

\subsection{Discriminability Analysis}\label{main-f-dis}
The design of an effective detector ${\cal D}$ relies heavily on a discriminative decomposition ${\cal F}$ \emph{w.r.t.} ${\bf{x}}$ and ${\bf{\delta}}$ in ${\bf{x'}}$. Such decomposition can be formulated as a mapping  ${\cal F}:{\cal X} \to {\cal C}$, where ${\cal X}$ is the image space and ${\cal C}$ is the space of decomposition coefficients.

To achieve a high discriminability, two constraints are typically imposed on the explicit forms of ${\cal F}$.

\begin{tcolorbox}
	[
	breakable,		                    
	colback= cyan!10!white,		            
	arc=0mm, auto outer arc,            
	boxrule= 0pt,                        
	boxsep = 0mm,                       
	left = 1mm, right = 1mm, top = 1mm, bottom = 1mm, 
	]
	{
		\begin{formulation} \textbf{(Distributive Property of Addition).} ${\cal F}$ should be distributed over addition to fulfill: ${\cal F}({\bf{x'}}) = {\cal F}({\bf{x}} + {\bf{\delta }}) = {\cal F}({\bf{x}}) + {\cal F}({\bf{\delta }})$, \emph{i.e.}, the decomposition of the adversarial example is equivalent to the sum of the decomposition of the clean example and the perturbation.
		\end{formulation}
	}
\end{tcolorbox}

The distributive property of addition facilitates the separation of natural-artificial data, and the operations like filtering \cite{ref23}, convolution \cite{ref24}, and inner product \cite{ref26}, \cite{ref27} from successful detectors satisfy this property.

\begin{tcolorbox}
	[
	breakable,		                    
	colback= cyan!10!white,		            
	arc=0mm, auto outer arc,            
	boxrule= 0pt,                        
	boxsep = 0mm,                       
	left = 1mm, right = 1mm, top = 1mm, bottom = 1mm, 
	]
	{
		\begin{formulation} \textbf{(Statistical Regularity).} ${\cal F}({\bf{x}})$ is expected to exhibit a consistent statistical pattern in ${\cal C}$ for any clean image ${\bf{x}} \in {\cal X}$, and ${\cal F}({\bf{\delta}})$ is also expected to exhibit another consistent statistical pattern in ${\cal C}$ for any ${\bf{\delta }}$ by pertinent ${\cal A}$; meanwhile, such two statistical patterns should be significantly different.
		\end{formulation}
	}
\end{tcolorbox}	

Here, we provide only a general description for the regularity. However, more specific assumptions about statistical regularity are needed in order to design ${\cal F}$ explicitly.

Next, we introduce two conjectures on the frequency and spatial distribution patterns of natural images and adversarial perturbations, as the main assumptions of our work. Note that such intuitive conjectures have general recognition in the research community \cite{ref23, ref24, ref27}, despite the lack of serious proofs.

\begin{tcolorbox}
	[
	breakable,		                    
	colback= yellow!10!white,		            
	arc=0mm, auto outer arc,            
	boxrule= 0pt,                        
	boxsep = 0mm,                       
	left = 1mm, right = 1mm, top = 1mm, bottom = 1mm, 
	]
	{
		\begin{conjecture} \textbf{(Frequency Pattern).} Natural images and adversarial perturbations have different frequency distributions. For natural images, their local smoothness and nonlocal self-similarity lead to the dominance of low-frequency components, with the \textbf{power law} of frequency-domain energy \cite{ref45}. For adversarial perturbations, the distribution of their frequency-domain energies very likely does not obey the same power law, since there is no term in the typical generation to directly control the frequency distribution of the perturbations \cite{ref76, ref77}.
		\end{conjecture}
	}
\end{tcolorbox}

\begin{tcolorbox}
	[
	breakable,		                    
	colback= yellow!10!white,		            
	arc=0mm, auto outer arc,            
	boxrule= 0pt,                        
	boxsep = 0mm,                       
	left = 1mm, right = 1mm, top = 1mm, bottom = 1mm, 
	]
	{
		\begin{conjecture} \textbf{(Spatial Pattern).} Natural images and adversarial perturbations have different spatial distributions. For natural images, each of their frequency components is distributed in spatial plane with \textbf{nonrandom structures} \cite{ref75}. For adversarial perturbations, the distribution of their frequency components very likely does not obey the same structure, since there is no term in the typical generation to directly control the spatial distribution of the perturbations \cite{ref76, ref77}.
		\end{conjecture}
	}
\end{tcolorbox} 

\emph{Remark.}  With Conjectures 1 and 2, we note that both frequency and spatial properties are of interest in the design of ${\cal F}$. More specifically, discriminative features should be formed in such frequency and spatial ranges where the distribution patterns of ${\bf{x}}$ and ${\bf{\delta}}$ differ, \emph{e.g.}, a naturally smooth region but with artificial high-frequency perturbations. Therefore, ${\cal F}$ is expected to analyze above frequency and spatial differences with sufficient resolution.

However, it is also well known that there is a trade-off between frequency and spatial resolutions of the orthogonal transform. In the related works, global transforms such as DST \cite{ref27} bias towards frequency resolution, at the cost of spatial resolution. As the opposite, local transforms such as DWT \cite{ref27} bias towards spatial resolution, at the cost of frequency resolution. Therefore, neither can provide rich spatial-frequency information.

Motivated by above facts, we introduce a mid-scale representation based on Krawtchouk polynomials, which provides a good trade-off between spatial and frequency resolutions than global/local transforms.
	
\subsection{Security Analysis}\label{main-f-sec}	 
Suppose the adversarial perturbations generated by (1) and (2) have a strong response (with main energy) on a subset of coefficients: ${{\cal C}_A} \subset {\cal C}$, and the detector-interested (with higher weights) subset of coefficients is denoted as ${{\cal C}_D} \subset {\cal C}$. Therefore, the effectiveness of the detector ${\cal D}$ is in fact built on the intersection ${{\cal C}_A} \cap {{\cal C}_D}$, where we denote the corresponding coefficient subset for a perturbation ${\bf{\delta }}$ as ${{\cal F}_{{{\cal C}_A} \cap {{\cal C}_D}}}({\bf{\delta }})$. 

\begin{tcolorbox}
	[
	breakable,		                    
	colback= cyan!10!white,		            
	arc=0mm, auto outer arc,            
	boxrule= 0pt,                        
	boxsep = 0mm,                       
	left = 1mm, right = 1mm, top = 1mm, bottom = 1mm, 
	]
	{
		\begin{formulation} \textbf{(Defense-aware Attack: General Objective).} With the above assumptions and notations, the objective of the defense-aware attack can be modeled on the correlation $\rho $ as
			\begin{equation}
				\rho ({{\cal F}_{{{\cal C}_A} \cap {{\cal C}_D}}}({{\bf{\delta }}_{{\rm{old}}}}),{{\cal F}_{{{\cal C}_A} \cap {{\cal C}_D}}}({{\bf{\delta }}_{{\rm{new}}}})) < \eta,
			\end{equation}
			for an image ${\bf{x}}$, with also objectives (1) and (2), where ${{\bf{\delta }}_{{\rm{old}}}}$ is the generated perturbation in the defense-unaware scenario, \emph{i.e.}, by only (1) and (2), and ${{\bf{\delta }}_{{\rm{new}}}}$ is the perturbation being generated.
		\end{formulation}
	}
\end{tcolorbox} 

\begin{tcolorbox}
	[
	breakable,		                    
	colback= cyan!10!white,		            
	arc=0mm, auto outer arc,            
	boxrule= 0pt,                        
	boxsep = 0mm,                       
	left = 1mm, right = 1mm, top = 1mm, bottom = 1mm, 
	]
	{
		\begin{formulation} \textbf{(Defense-aware Attack: A Special Case).} In practice, objective (3) can be converted into another easily implemented objective — directly shifting the main energy of the perturbation ${{\bf{\delta }}_{{\rm{new}}}}$ out of ${{\cal C}_A} \cap {{\cal C}_D}$:
			\begin{equation}
				{\rm{||}}{{\cal F}_{{{\cal C}_A} \cap {{\cal C}_D}}}({{\bf{\delta }}_{{\rm{new}}}}){\rm{||}} < \lambda,
			\end{equation}
		such an objective allows defense-aware perturbations to form on the relative complement ${\cal C}\backslash ({{\cal C}_A} \cap {{\cal C}_D})$, hence destroying the consistent pattern of ${\cal F}({{\bf{\delta }}_{{\rm{old}}}})$ and ${\cal F}({{\bf{\delta }}_{{\rm{new}}}})$ on the adversarial and detector-interested ${{\cal C}_A} \cap {{\cal C}_D}$.
		\end{formulation}
	}
\end{tcolorbox} 

\emph{Remark.}  Note that the above new objectives in the defense-aware scenario, \emph{i.e.}, (3) or (4), rely in fact on the sufficient knowledge of features ${{\cal C}_A} \cap {{\cal C}_D}$. For the related works with few fixed features \cite{ref23, ref24}, the adversary is able to easily identify the critical ${{\cal C}_A} \cap {{\cal C}_D}$ and hence successfully evade the detector ${\cal D}$ as well as fool the model ${\cal M}$ by the objectives (1) and (2) with (3) or (4).

Motivated by above facts, we introduce a secrecy strategy on ${\cal F}$ by random feature selection with keys, which confuses the boundary between to-be-attacked features ${\cal C}\backslash ({{\cal C}_A} \cap {{\cal C}_D})$ and to-be-evaded features ${{\cal C}_A} \cap {{\cal C}_D}$.

\begin{figure*}[!t]
	\centering
	\subfigure[$P = 0.25$]{\includegraphics[scale=0.8]{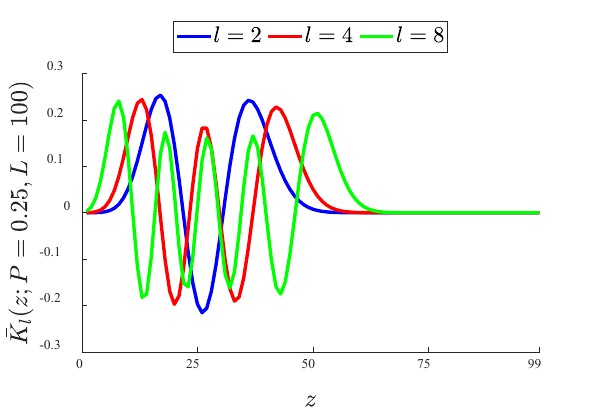}}
	\subfigure[$P = 0.5$]{\includegraphics[scale=0.8]{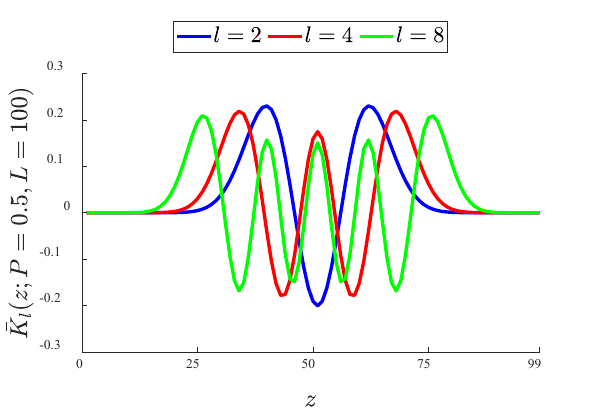}}
	\subfigure[$P = 0.75$]{\includegraphics[scale=0.8]{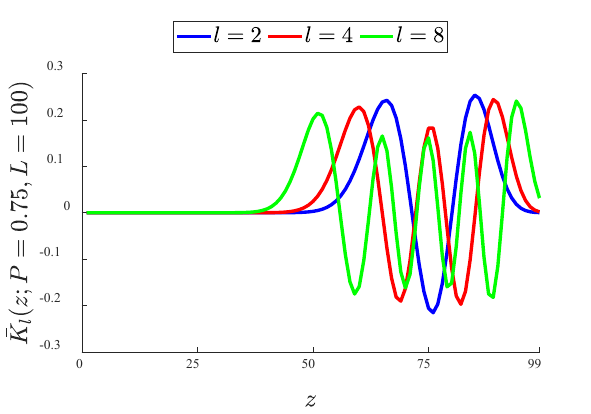}}
	\centering
	\caption{Illustration for the weighted Krawtchouk polynomials ${\bar K_l}(z;P,L)$ with $l = \{ 2,4,8\} $, $P = \{ 0.25,0.5,0.75\}$, and $L = 100$. Note that the number and location of zeros of $\bar K$ can be adjusted explicitly by $l$ and $P$, respectively, meaning the time-frequency discriminability of the represented image information.}
\end{figure*}

\section{Towards Accurate and Secure Detector: Methodology}\label{main-m}
In this section, we specify the proposed detector against adversarial perturbation. We will first give an overview drawing a high-level intuition for readers, and then the main techniques within the methodology are presented separately.

\subsection{Overview}\label{main-m-over}
In general, the proposed detector consists of three main steps, going through training and inference phases.

As shown in Fig. 1, the training of our detector aims to fit a mapping from a set of adversarial/clean image examples to the corresponding labels. For an efficient mapping, our detector is equipped with three steps that perform decomposition, featuring, and classification, respectively.

\begin{itemize}
	\item Regarding the \emph{decomposition}, the image is projected into a space defined by Krawtchouk polynomials, in which the clean image ${\bf{x}}$ and adversarial perturbation ${\bf{\delta }}$ are better separable from ${\bf{x'}}$ (see also the discriminability analysis in Section \ref{main-f-dis}). As for security, the frequency parameters $(n,m)$ and spatial parameters $({P_x},{P_y})$ are determined by the host key, which is considered as an opaque mechanism for decomposition (see also the security analysis in Section \ref{main-f-sec}). The description on decomposition will be presented in Section \ref{main-m-decomp}.
	\item Regarding the \emph{featuring}, a compact-but-expressive feature vector is formed on the obtained decomposition coefficients. Here, the statistical regularity in the frequency and spatial domains are introduced, where coefficients are integrated and enhanced as features by means of above beneficial priors. The description on featuring will be presented in Section \ref{main-m-feat}.
	\item Regarding the \emph{classification}, the above features are fed into a Support Vector Machine (SVM) for an automatic two-class separation of feature space. Note that the decomposition and featuring are non-learning, and the SVM is the only learning part of the detector. The description on classification will be presented in Section \ref{main-m-svm}.
\end{itemize}

As shown in Fig. 2, the trained detector is an accurate and secure defense tool against adversarial attacks, technically enabled by a spatial-frequency discriminative decomposition with secret keys. It can be deployed in various real-world scenarios, where a DLaaS scenario is chosen as an example. For an image under analysis, our detector (with the same key and SVM parameters in training) predicts whether it contains adversarial perturbations. By such prediction, the DLaaS is able to deny the service when the adversarial perturbation is revealed.

\subsection{Spatial-frequency Discriminative Decomposition with Secret Keys}\label{main-m-decomp}
Our defense framework starts with a spatial-frequency discriminative decomposition of input examples. In this subsection, we discuss the explicit definition of such a decomposition, as well as the security enhancement strategy with secret keys.

\begin{tcolorbox}
	[
	breakable,		                    
	colback= orange!10!white,		            
	arc=0mm, auto outer arc,            
	boxrule= 0pt,                        
	boxsep = 0mm,                       
	left = 1mm, right = 1mm, top = 1mm, bottom = 1mm, 
	]
	{
		\begin{definition} \textbf{(Orthogonal Decomposition).} The orthogonal decomposition of an image function $f \in {\cal X}$, denoted as ${\cal F}$, is defined as the inner product of the image function $f$ and the basis function $V$ \cite{ref46}:
		\begin{equation}
			{\cal F}(f) = \iint_{D}{V_{nm}^*(x,y)f(x,y)dxdy},
		\end{equation}
		where the frequency parameters $(n,m) \in {\mathbb{Z}^2}$, the domain of basis function $D \subset \{ (x,y) \in {\mathbb{R}^2}\}$, the asterisk $ * $ denotes the complex conjugate. Here, the basis function $V$ satisfy orthogonality over the domain $D$ as
		\begin{equation}
			\iint_{D}{{V_{nm}}(x,y)V_{n'm'}^*(x,y)dxdy} = {\delta _{nn'}}{\delta _{mm'}},
		\end{equation}
		where $\delta $ is the Kronecker delta function: ${\delta _{ab}} = [a = b]$.
		\end{definition}
	}
\end{tcolorbox}

With Definition 1, one can note that the orthogonal decomposition methods used in existing detectors all have the form (5) and (6), and their difference lies in the definition of the basis function $V$. For example, the DST and DWT in the detector of Agarwal \emph{et al.} \cite{ref27} are with the global trigonometric basis and local wavelet basis, respectively.

In this paper, we define a mid-scale basis by Krawtchouk polynomials for a better trade-off between spatial and frequency resolutions. In \cite{ref47}, Yap \emph{et al.} introduced a new set of decompositions from the weighted Krawtchouk polynomials. Here, the orthogonality of the weighted Krawtchouk polynomials ensures information preservation and decorrelation; also no numerical approximation is involved, since the weighted Krawtchouk polynomials are discrete like digital images. Such properties are what make it a good choice for this paper.

\begin{tcolorbox}
	[
	breakable,		                    
	colback= orange!10!white,		            
	arc=0mm, auto outer arc,            
	boxrule= 0pt,                        
	boxsep = 0mm,                       
	left = 1mm, right = 1mm, top = 1mm, bottom = 1mm, 
	]
	{
		\begin{definition} \textbf{(Krawtchouk Basis).} The Krawtchouk basis function $V$  \cite{ref47} is defined as
			\begin{equation}
				{V_{nm}}(x,y) = {\bar K_n}(x;{P_x},W){\bar K_m}(y;{P_y},H),
			\end{equation}
		where domain of basis function $D = \{ (x,y) \in [0,1,...,W] \times [0,1,...,H]\}$ with image size $W \times H$, frequency parameters $(n,m) \in [0,1,...,W] \times [0,1,...,H]$, spatial parameters $({P_x},{P_y}) \in {(0,1)^2}$, and weighted Krawtchouk polynomials $\bar K$ are defined as
   			 \begin{equation}
   			 	\begin{split}
   			 		{\bar K_l}(z;P,L) = 
   			 		&\sqrt {\frac{{\left( {\begin{array}{*{20}{c}}
   			 							L\\
   			 							z
   		 							\end{array}} \right){P^z}{{(1 - P)}^{L - z}}}}{{{{( - 1)}^l}{{(\frac{{1 - P}}{P})}^l}\frac{{l!\Gamma ( - L)}}{{\Gamma (l - L)}}}}} \\
   	 							&\cdot {}_2{F_1}( - l, - z; - L;\frac{1}{P}),
    			\end{split} 
    		\end{equation}
    	where hypergeometric function ${}_2{F_1}$ is defined as
			\begin{equation}
				{}_2{F_1}(a,b;c;d) = \sum\limits_{k = 0}^\infty  {\frac{{{{(a)}_k}{{(b)}_k}}}{{{{(c)}_k}}}} \frac{{{d^k}}}{{k!}},
			\end{equation}
	    with Pochhammer symbol: ${(a)_k} = \Gamma (a + k)/\Gamma (a)$. For more efficient computation of (8), \emph{i.e.}, avoiding the infinite summation in (9), we introduce the recursive formula for weighted Krawtchouk polynomials $\bar K$:
	    	\begin{equation}
				\begin{split}
					{\bar K_{l + 1}} = &\frac{{\sqrt {\frac{{(1 - P)(l + 1)}}{{P(L - l)}}} (LP - 2lP + l - z){{\bar K}_l}}}{{P(l - L)}} \\
					& - \frac{\sqrt {\frac{{{{(1 - P)}^2}(l + 1)l}}{{{P^2}(L - l)(L - l + 1)}}} l(1 - P){{\bar K}_{l - 1}}}{{P(l - L)}},
				\end{split}		
			\end{equation}
		with initial items ${\bar K_1}$ and ${\bar K_0}$:
			\begin{equation}
				{\bar K_1}(z;P,L) = (1 - \frac{z}{{PL}}){\bar K_0},
			\end{equation}
			\begin{equation}
				{\bar K_0}(z;P,L) = \sqrt {\left( {\begin{array}{*{20}{c}}
					L\\
					z
				\end{array}} \right){P^z}{{(1 - P)}^{L - z}}}.
			\end{equation}
		\end{definition}
	}
\end{tcolorbox}

By substituting the basis of Definition 2 into the decomposition of Definition 1, we have formulated the Krawtchouk decomposition, which is fundamental in our detector.

\subsubsection{Discriminability Analysis}\label{main-m-decomp-dis}

Next, we will discuss the key property of Krawtchouk decomposition, \emph{i.e.}, time-frequency discriminability, and its role in the detection of adversarial perturbation.

\begin{tcolorbox}
	[
	breakable,		                    
	colback= purple!8!white,		            
	arc=0mm, auto outer arc,            
	boxrule= 0pt,                       
	boxsep = 0mm,                       
	left = 1mm, right = 1mm, top = 1mm, bottom = 1mm, 
	]
	{
		\begin{property} \textbf{(Time-frequency Discriminability).} The frequency and spatial properties of the represented image information by Krawtchouk decomposition can be controlled with the frequency parameters $(n,m)$ and spatial parameters $({P_x},{P_y})$, respectively \cite{ref47}.
		\end{property}
	}
\end{tcolorbox}

\emph{Remark.} In the study of image representation, it has been found that the frequency and spatial properties of orthogonal decomposition rely on the number and location of zeros of the basis functions, respectively \cite{ref48}. Specific to this paper, the core of time-frequency discriminability in Krawtchouk decomposition is that the number and location of zeros can be adjusted explicitly by $(n,m)$ and $({P_x},{P_y})$, respectively:

\begin{itemize}
	\item The number of zeros of the 1D ${\bar K_l}(z;P,L)$ is proportional to $l$. As for the 2D ${V_{nm}}(x,y) = {\bar K_n}(x){\bar K_m}(y)$, the similar conclusion holds \emph{w.r.t.} the $n$ and $m$ at the $x$-direction and $y$-direction, respectively..
	\item The location of zeros of the 1D ${\bar K_l}(z;P,L)$ is biased towards 0 when $P < 0.5$, uniform when $P = 0.5$, and biased towards 1 when $P > 0.5$, where the more deviation of $P$ from 0.5 is, the more biased the distribution of zeros is. As for the 2D ${V_{nm}}(x,y) = {\bar K_n}(x){\bar K_m}(y)$, the similar conclusion holds \emph{w.r.t.} the ${P_x}$ and ${P_y}$ at the $x$-direction and $y$-direction, respectively. 
\end{itemize}

In Fig. 3, we illustrate 1D plots of weighted Krawtchouk polynomials ${\bar K_l}(z;P,L)$ for a high-level intuition of such time-frequency discriminability. Here, the plots under different parameter settings: $l = \{ 2,4,8\} $ and $P = \{ 0.25,0.5,0.75\}$ with $L = 100$. As can be expected, changing $l$ will change the number of zeros of ${\bar K}$, which in turn corresponds to a change in the frequency properties. As for $P$, the change of its value will change the distribution of zeros, which in turn corresponds to a change in the spatial properties. The 2D plots of ${V_{nm}}(x,y)$ \emph{w.r.t.} $(n,m)$ and $({P_x},{P_y})$ are given in Fig. 1, where frequency and spatial property changes at the $x$ and $y$ directions can be observed.

\begin{tcolorbox}
	[
	breakable,		                    
	drop shadow southeast, enhanced,    
	colback= OliveGreen!7!white,		        
	colframe=OliveGreen!50!white,				    	
	arc=2mm, auto outer arc,            
	boxrule=1pt,                        
	boxsep = 0mm,                       
	left = 1mm, right = 1mm, top = 1mm, bottom = 1mm, 
	]	
	{
		\begin{main result} \textbf{(Increased Discriminability for Adversarial Perturbation).}
			The Krawtchouk decomposition is  discriminative for adversarial perturbation due to the following factors: 1) The Krawtchouk decomposition defined by the inner product (Definitions 1 and 2) satisfies the distributive property of addition (Formulation 1); 2) The Krawtchouk decomposition with time-frequency discriminability (Property 1) is able to explore the statistical regularity (Formulation 2), when the frequency pattern (Conjecture 1) and spatial pattern (Conjecture 2) hold in natural images and adversarial perturbations.
		\end{main result}
	}
\end{tcolorbox}

\emph{Remark.} Although decompositions in competing detectors, \emph{e.g.}, denoising filters \cite{ref23}, SRM \cite{ref24}, DCT \cite{ref26}, DST \cite{ref27}, and DWT \cite{ref27}, all satisfy the distributive property of addition (Formulation 1), achieving the statistical regularity (Formulation 2) is still an open problem. Global decompositions, \emph{e.g.}, DST, are biased towards frequency resolution, thus better exploiting frequency patterns, but fail in mining spatial patterns. In contrast, local decompositions, \emph{e.g.}, denoising filters, SRM, and DWT, can fully exploit the spatial patterns, but only provide limited frequency resolution and thus fail in mining frequency patterns. As a mid-scale representation, the Krawtchouk decomposition provides a good trade-off between spatial and frequency resolutions than above global/local methods. It is thus expected to reveal a more comprehensive pattern of adversarial perturbation over the both spatial and frequency dimensions.

\subsubsection{Security Analysis}\label{main-m-decomp-sec}

With the Property 1 and the Main Result 1, we have analyzed the theoretical effectiveness of the Krawtchouk decomposition in detecting adversarial perturbations. Next, we will discuss the secrecy strategy of decomposition to provide certain security guarantees against defense-aware attacks.

In the implementation of previous detectors, the parameters of the orthogonal decomposition were often determined \emph{explicitly} by the host. In our implementation, they are determined by a pseudorandom number generator, \emph{i.e.}, random feature selection \cite{ref31}. The host keeps the seed value of the generator secret, where such a seed is considered as the key for decomposition. In fact, the secret key determines the set of coefficients (\emph{w.r.t.} spatial and frequency features) that the detector can explore (see also Property 1).

\begin{tcolorbox}
	[
	breakable,		                    
	drop shadow southeast, enhanced,    
	colback= OliveGreen!7!white,		        
	colframe=OliveGreen!50!white,				    	
	arc=2mm, auto outer arc,            
	boxrule=1pt,                        
	boxsep = 0mm,                       
	left = 1mm, right = 1mm, top = 1mm, bottom = 1mm, 
	]	
	{
		\begin{main result} \textbf{(Increased Security for Defense-aware Attack).}
			The random feature selection increase the security for the defense-aware attack due to the following factor. After the randomization of Krawtchouk decomposition, the adversary is more difficult to identify the adversarial and detector-interested ${{\cal C}_A} \cap {{\cal C}_D}$ (Formulations 3 and 4). Furthermore, such secrecy strategy confuses the boundary between to-be-attacked features ${\cal C}\backslash ({{\cal C}_A} \cap {{\cal C}_D})$ and to-be-evaded features ${{\cal C}_A} \cap {{\cal C}_D}$, resulting in a dilemma for the adversary in attacking the model ${\cal M}$ and evading the detector ${\cal D}$.
		\end{main result}
	}
\end{tcolorbox}

\emph{Remark}. Some decompositions in detectors, \emph{e.g.}, denoising filters \cite{ref23} and SRM \cite{ref24}, are based on few fixed filters/bases. Such design is easy for the adversary to form an effective evasion of the detector. Other detectors use a more comprehensive bases as the decomposition, \emph{e.g.}, DCT \cite{ref26}, DST \cite{ref27}, and DWT \cite{ref27}, which enhances both the discriminability and security. However, the risk of evading the detector remains, where the adversary can still determine the critical ${{\cal C}_A} \cap {{\cal C}_D}$ due to the transparency in the design of detector and the definition of decomposition. In our work, the random feature selection provides a secrecy mechanism, placing fundamental dilemma for the defense-aware attack.

We would like to state that the Main Result 2 is not perfectly secure — it should be considered as a means of increasing the attack difficulty (also security) than the few fixed features of many existing detectors: 1) Note that our security is a special case of the general modeling by Chen \emph{et al.} \cite{ref31} for the random feature selection in forensic detectors. Here, such random feature selection permits to greatly enhance security against full-feature attack (\emph{i.e.}, the worst case in \emph{one-shot} attacks), according to the detailed proofs and numerical results of \cite{ref31}. 2) For \emph{multi-shot} attacks, \emph{e.g.} model reverse engineering, they can indeed theoretically successfully attack any detector with random feature selection. On the other hand, their black-box mechanism forces them to rely on many queries about the inputs and outputs of the detector. In practice, this short-term over-querying behavior is easily identified and blocked, and has become a current consensus in server security (like DDoS mitigation). In addition, changing the host key periodically is a practical trick that can further increase difficulty (also security) \emph{w.r.t.} multi-shot attacks.

\subsection{Frequency-band Integration and Feature Enhancement}\label{main-m-feat}
We first recall Section \ref{main-m-decomp} and provide some proper notations.

After the Krawtchouk decomposition, the set of spatial-frequency discriminability (Main Result 1) coefficients is denoted as ${\cal C} = \{ {c_{n,m,{P_x},{P_y}}}\}$, where the random sampling provides certain security guarantees \emph{w.r.t.} defense-aware attacks (Main Result 2). There are some guidelines on the sampling strategy of the parameters: 1) a very comprehensive set of candidate parameters is set first, motivated by both the accuracy and security of the detector; 2) some parameters in this set are then blocked randomly with a certain probability, motivated by the security of the detector, where trade-off between accuracy and security can be controlled by the blocking probability — higher blocking probability leads to potentially higher security but lower accuracy, and vice versa (see also proofs by Chen \emph{et al.} \cite{ref31}); 3) the final set of parameters is formed by further automatic selection or learning, where more discriminative parameters (features) under a given training set can be automatically filtered by statistical correlation or more complex feature selection learning.

\emph{Frequency-band Integration.} With a given sampling strategy of the parameters, the coefficient set ${\cal C}$ is very high-dimensional and exhibits certain information redundancy, where direct knowledge engineering is usually inefficient. Here, the statistical regularity (Formulation 2) are explored for forming a compact-but-expressive feature vector ${\cal V}$ on the coefficient set ${\cal C}$. More specifically, the coefficients of similar frequency properties in ${\cal C}$ are integrated as a component of feature vector ${\cal V}$, inspired by the frequency pattern (Conjecture 1) and spatial pattern (Conjecture 2).
First, the space of frequency coefficients of the Krawtchouk decomposition, \emph{i.e.}, $(n,m) \in [0,1,...,W] \times [0,1,...,H]$, is divided equally into ${\# _B}$ bands under ${\ell _2}$ norm
\begin{equation}
	\begin{split}
		&{{\cal B}_i} = \{ (n,m): \frac{{i - 1}}{{{\# _B}}}||(W,H)|{|_2} \le ||(n,m)|{|_2}\\
		& \quad \quad \quad < \frac{i}{{{\# _B}}}||(W,H)|{|_2}\},
	\end{split}
\end{equation}
where $i = 1,2,...,{\# _B}$ and ${\cal C}  =  \cup _{i = 1}^{{\# _B}}{{\cal B}_i}$.

Then, the coefficients ${c_{n,m,{P_x},{P_y}}}$ in the frequency band ${{\cal B}_i}$ are considered with similar frequency properties and they are integrated as a feature component
\begin{equation}
	{{\cal V}_i}({P_x},{P_y}) = \sum\limits_{(n,m) \in {{\cal B}_i}} {{c_{n,m,{P_x},{P_y}}}},
\end{equation}
where the feature vector ${\cal V}  = \{ {{\cal V}_i}({P_x},{P_y})\} $.

\emph{Feature Enhancement.} In fact, the feature ${\cal V}$ obtained by frequency-band integration can still be enhanced, \emph{i.e.}, improving compactness and expressiveness. Here, we present two simple enhancement strategies, note that they are optional and not mandatory in the implementation.

\begin{itemize}
	\item Weighting: Starting from pairs of clean and adversarial examples in the training set, we evaluate in which frequency bands the features exhibit stronger variability. Simple functions (\emph{e.g.} Gaussian functions) reflecting such variability can be set to weight the obtained feature vector, where the more discriminative bands are highlighted.
	\item Ranking: Starting from the features and labels of the training examples, we calculate the correlation of each feature dimension/component with the labels. Then, the feature vector is re-ranked according to the relevance, where the dimension/component with low relevance can be dropped directly.
\end{itemize}

\subsection{SVM Prediction}\label{main-m-svm}
After the featuring of Section \ref{main-m-feat}, we formed compact-but-expressive feature ${\cal V}$ from the Krawtchouk coefficient set ${\cal C}$. Considering the discriminative power of such features, a simple SVM classification model is sufficient to support a two-class separation of feature space. Since the SVM prediction is not our main technical contribution, only a brief description is provided.

Here, a set of paired clean-adversarial image examples is employed as the training set. The features formed on training set and corresponding labels are fed into a SVM. The training process attempt to achieve a proper mapping from features to labels by minimizing the loss. The learned SVM parameters are then saved as an important part of the detector, and such parameters will be shared to the inference phase of the practical deployment. Note that the SVM is the only learning part of the detector.

\begin{figure*}[!t]
	\centering
	\subfigure{\includegraphics[scale=0.9]{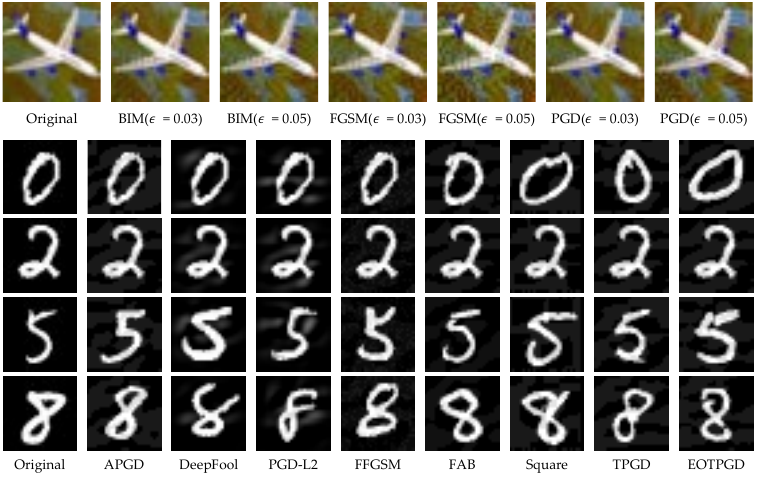}}
	\subfigure{\includegraphics[scale=0.85]{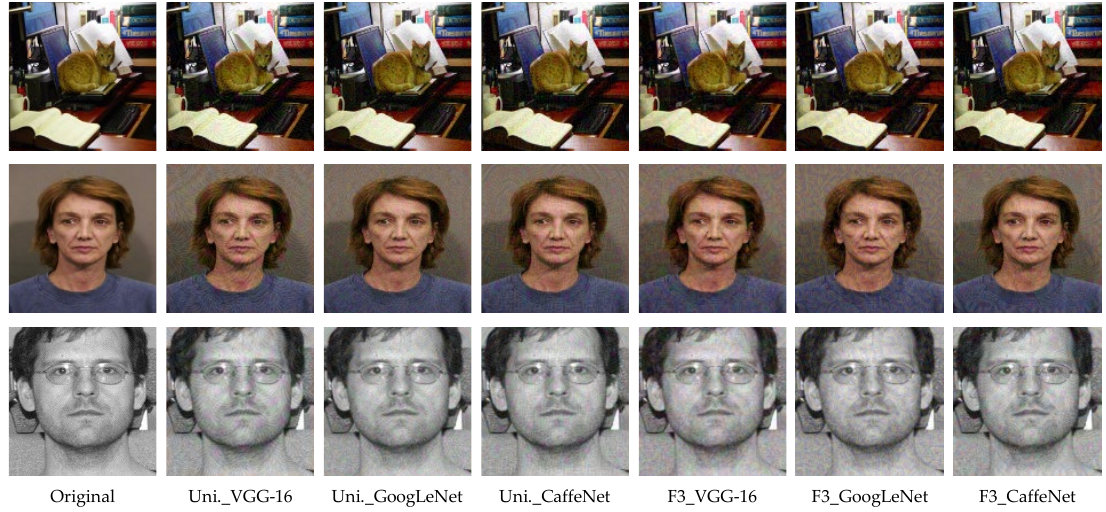}}
	\caption{Illustration for some datasets and adversarial attacks involved in experiments.}
\end{figure*}

\section{Experiments}\label{exp}

In this section, we will evaluate the capability of the proposed method to detect adversarial examples by extensive quantitative analysis. The code is available online at \texttt{https://github.com/ChaoWang1016/ASD}.

We first provide the basic setup \emph{w.r.t.} models, attacks, and defenses in the following experiments. Then, the proposed detector is evaluated with benchmarking, crossing, and challenging protocols, thus determining its position \emph{w.r.t.} current state-of-the-art detectors and its effectiveness for realistic scenarios. Finally, we discuss limitations regarding the defense scope of the proposed detector.

\subsection{Experiment Setup}\label{exp-set}
In general, the experiments of this paper involve the setup of three aspects: 1) the foundational deep models along with datasets, which are going to be attacked/defended; 2) the adversarial perturbation generators for attacking; 3) the adversarial perturbation detectors for defensing. In Fig. 4, we provide the visualization for some adversarial images.

\begin{figure}[!t]
	\centering
	\includegraphics[scale=0.5]{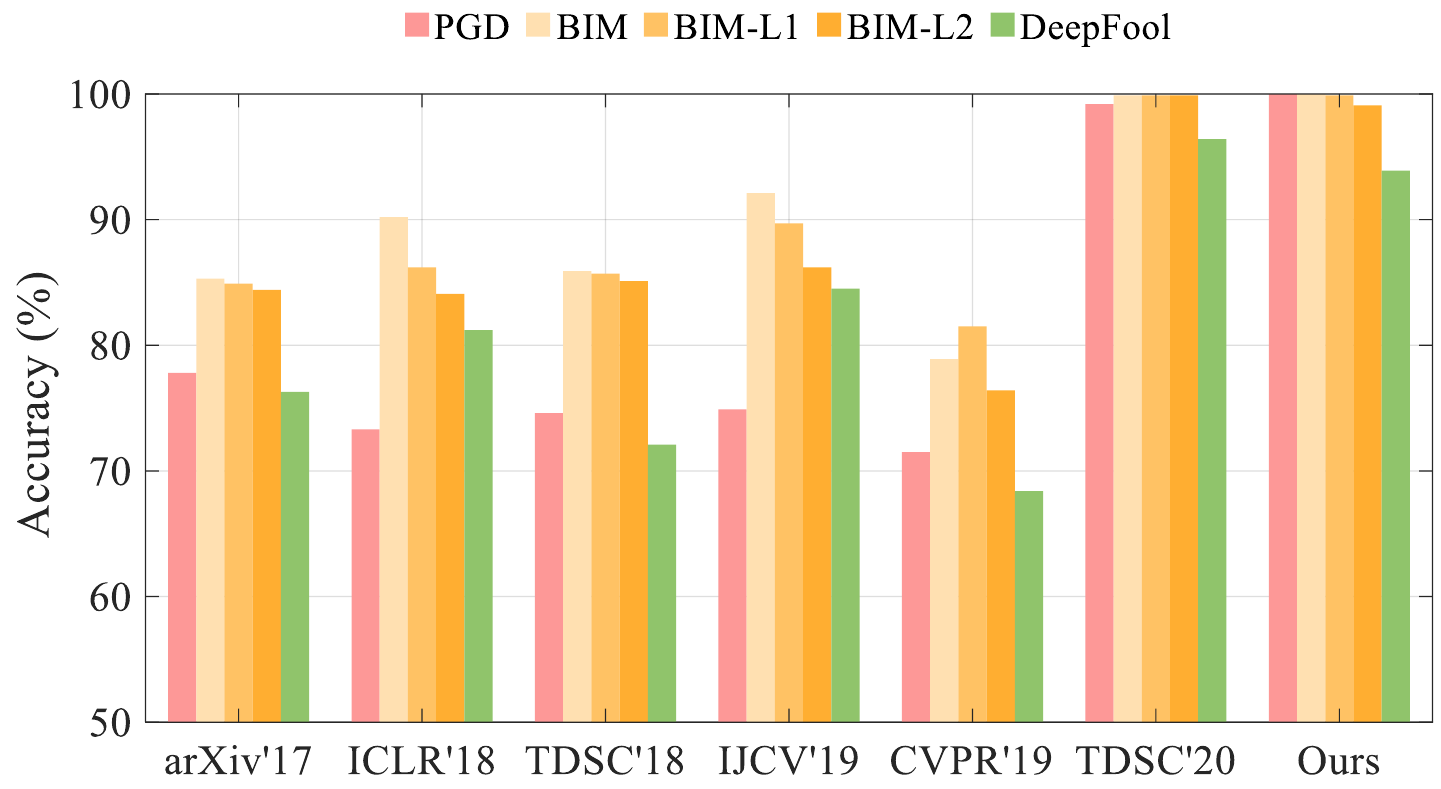}
	\centering
	\caption{The benchmarking of adversarial perturbation detection accuracy for different detectors on the MNIST dataset.}
\end{figure}

\begin{figure}[!t]
	\centering
	\includegraphics[scale=0.5]{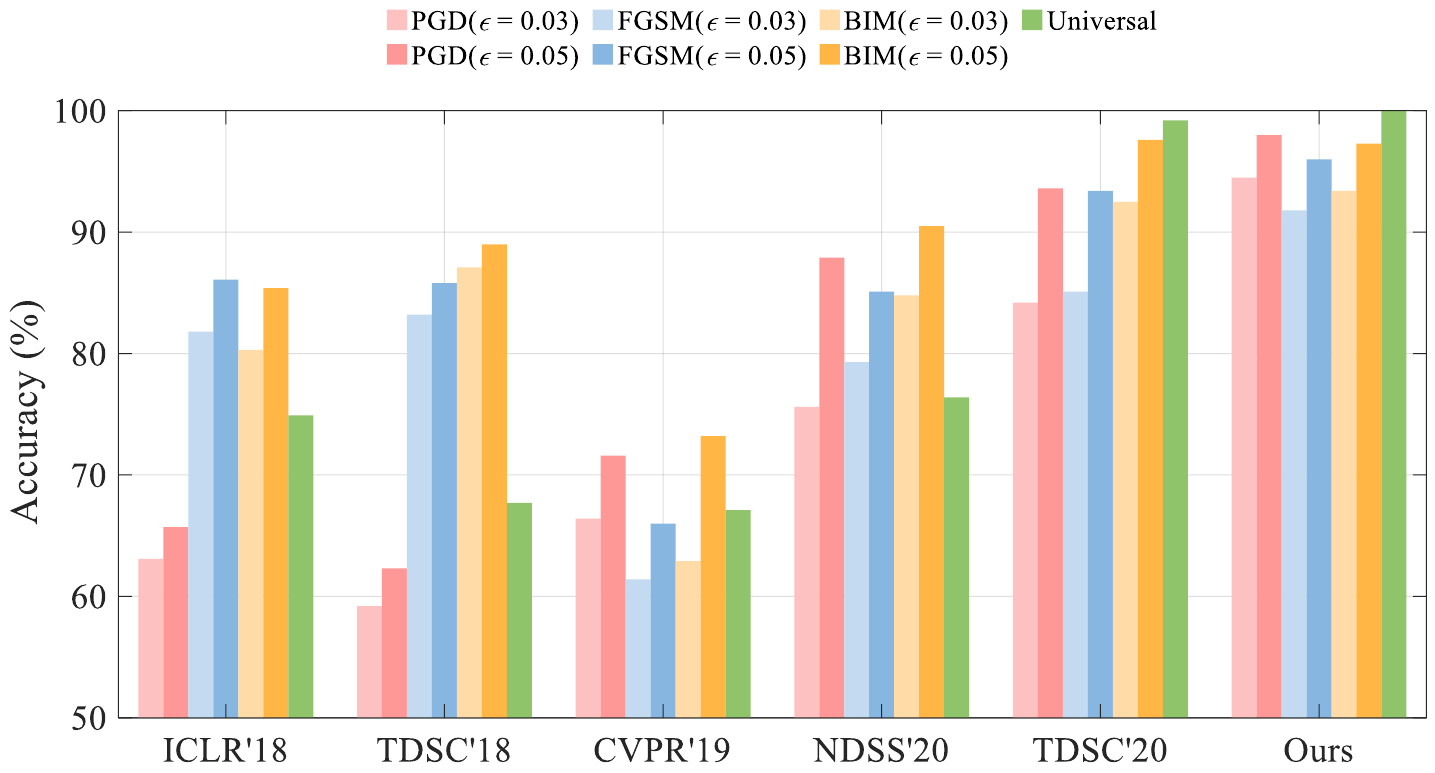}
	\centering
	\caption{The benchmarking of adversarial perturbation detection accuracy for different detectors on the CIFAR-10 dataset.}
\end{figure}

The deep models involved in experiments are LeNet \cite{ref49}, VGG-16 \cite{ref50}, GoogLeNet \cite{ref51}, CaffeNet \cite{ref52}, ResNet-50 \cite{ref85}, Vit-B/16 \cite{ref86}, and Swin-B \cite{ref87}. The datasets involved in experiments are MNIST \cite{ref53},  CIFAR-10 \cite{ref54},  MEDS \cite{ref55},  Multi-PIE \cite{ref56}, PubFig \cite{ref57}, and ImageNet \cite{ref58}. The adversarial attacks involved in experiments are  FGSM \cite{ref9},  BIM \cite{ref7}, PGD \cite{ref59}, APGD \cite{ref60}, DeepFool \cite{ref61}, FFGSM \cite{ref62}, FAB \cite{ref63}, Square \cite{ref64}, TPGD \cite{ref65}, EOTPGD \cite{ref66}, Universal \cite{ref11}, F3 \cite{ref67}, and Jitter \cite{ref88}.	

The adversarial perturbation detectors (\emph{i.e.}, comparison methods) involved in experiments are
\begin{itemize}
	\item arXiv’17 \cite{ref68} by looking at Bayesian uncertainty estimates;
	\item ICLR’18 \cite{ref69} by detecting out-of-distribution images;
	\item TDSC’18 \cite{ref23} by denoising filters;
	\item IJCV’19 \cite{ref70} by characterizing abnormal filter response behavior;
	\item CVPR’19 \cite{ref24} by steganalysis and spatial rich model;
	\item NDSS’19 \cite{ref71} by neural network invariant checking;
	\item TDSC’20 \cite{ref27} by global and local orthogonal decomposition.
\end{itemize}

\subsection{Benchmarking Experiments}\label{exp-bench}
In this part, we provide benchmarking evaluations of the proposed detector \emph{w.r.t.} current state-of-the-art detectors. Note that the comparison results for above methods are mainly cited from \cite{ref27}. As detailed in \cite{ref27}, we used the exact same protocol for the next benchmarking experiments to ensure a fair comparison.

\subsubsection{MNIST}\label{exp-bench-min}
Fig. 5 shows the accuracy comparison of 7 detectors \emph{w.r.t.} 7 attacks on the MNIST dataset. Here, 9000 clean images are selected from the MNIST, and then 9000 corresponding perturbed images are formed by each attack. For each competing detector, it is trained and tested based on the above images, with a training-testing split of 50\%-50\% on both original and perturbed images. It can be observed that both TDSC’20 and proposed detectors achieve significant gains in detection accuracy \emph{w.r.t.} other advanced detectors over different attacks. A possible explanation is that the orthogonal decomposition provides more comprehensive representations of perturbations for the classifier. Note that we complement more comprehensive results on MNIST with different attacks in Appendix B.

\begin{figure}[!t]
	\centering
	\includegraphics[scale=0.5]{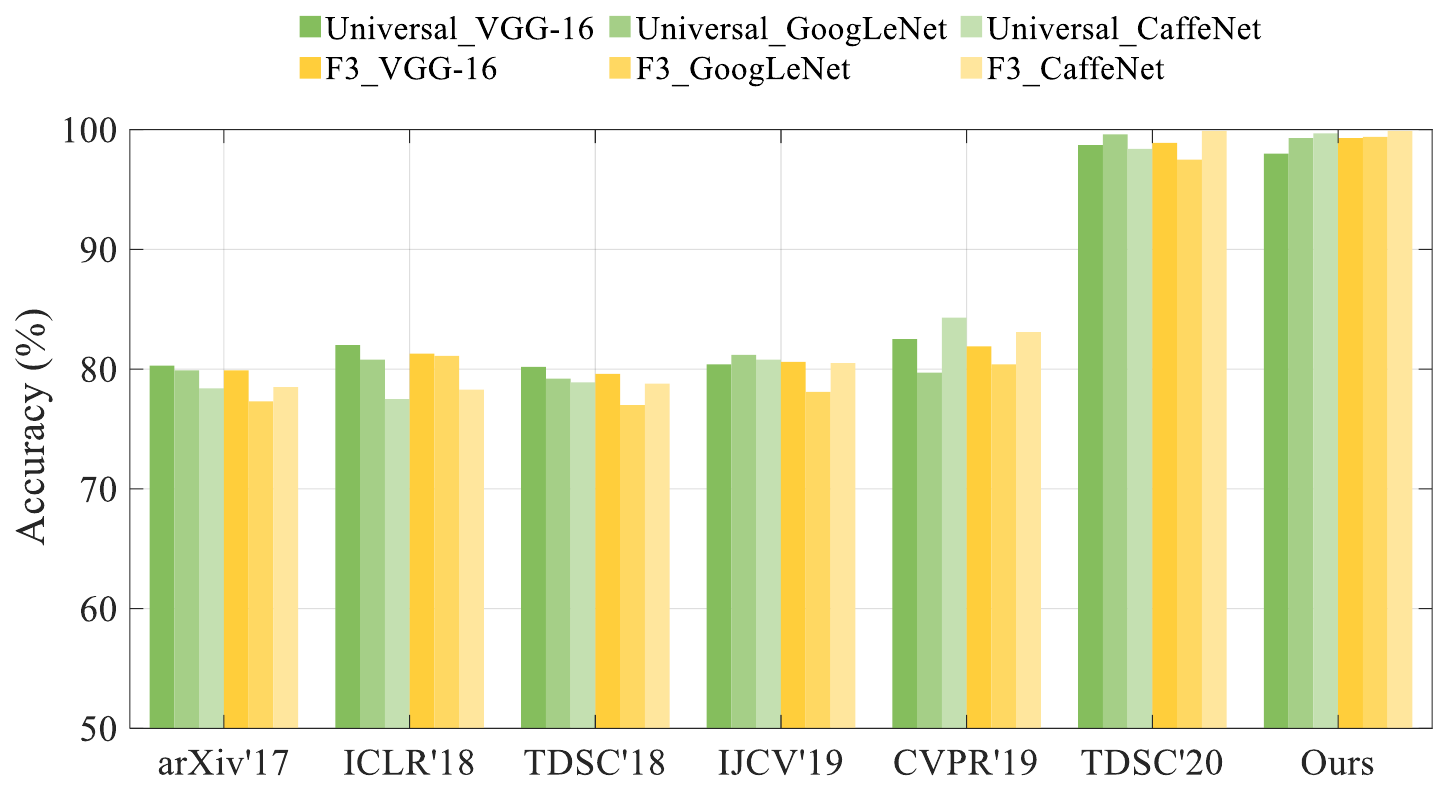}
	\centering
	\caption{The benchmarking of adversarial perturbation detection accuracy for different detectors on the MEDS dataset.}
\end{figure}

\begin{figure}[!t]
	\centering
	\includegraphics[scale=0.5]{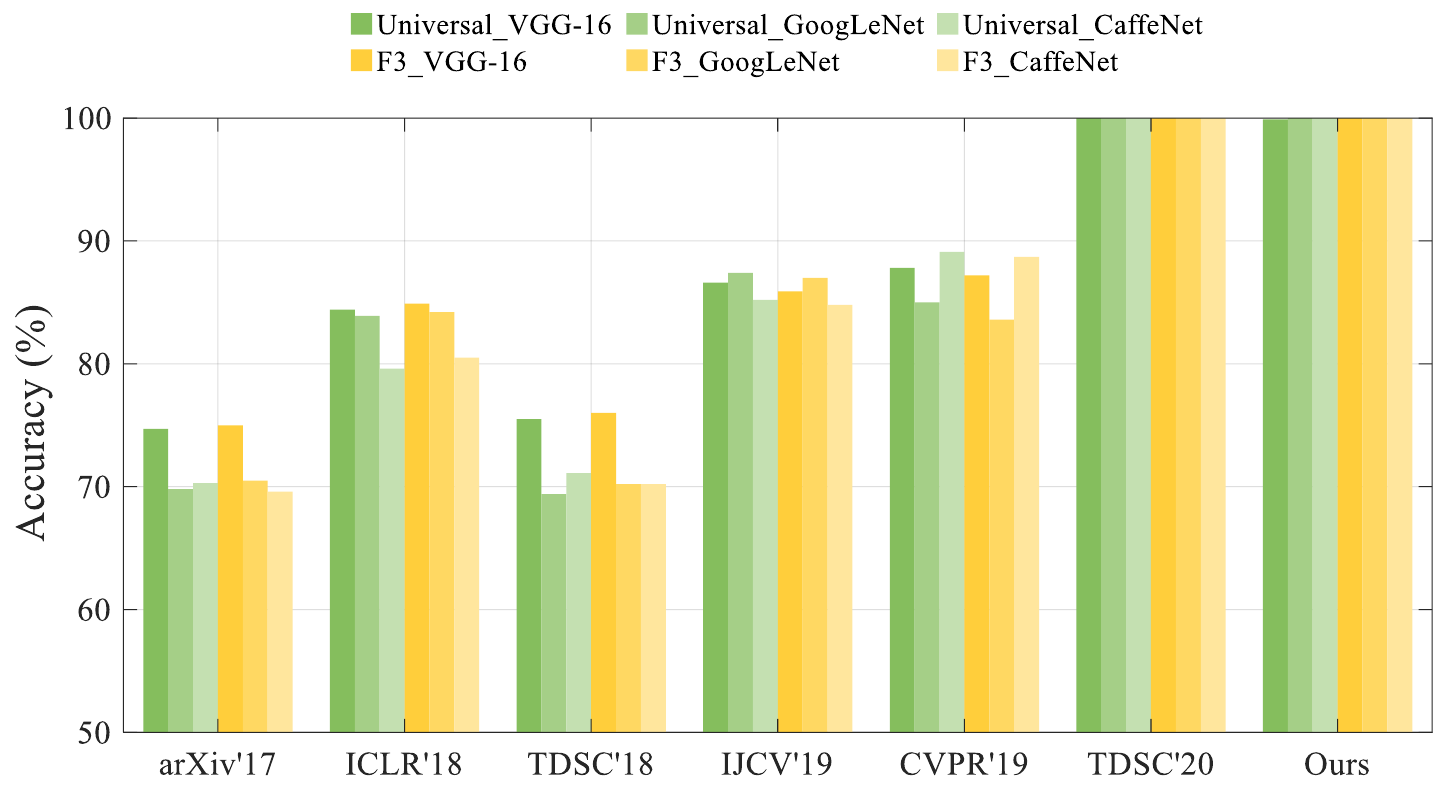}
	\centering
	\caption{The benchmarking of adversarial perturbation detection accuracy for different detectors on the Multi-PIE dataset.}
\end{figure}

\subsubsection{CIFAR-10}\label{exp-bench-cif}
Fig. 6 shows the accuracy comparison of 6 detectors \emph{w.r.t.} 7 attacks on the CIFAR-10 dataset. Here, the experiment covers 10000 clean images and 10000 corresponding perturbed images for each attack, also with 50\%-50\% training-testing split. Compared to the results on MNIST, one can note a significant performance degradation in competing methods, even $ > $ 10\% degradation in ICLR’18 and TDSC’18 \emph{w.r.t.} PGD attacks. This is mainly due to the richer patterns of image content in CIFAR-10, which acts as a strong interference for representing perturbation patterns. Among them, the proposed detector exhibits the least performance degradation over different attacks, even when compared to TDSC’20. Such a phenomenon further confirms the effectiveness of our spatial-frequency discriminative decomposition.

\subsubsection{MEDS and Multi-PIE}\label{exp-bench-face}
In Fig. 7 and Fig. 8, we provide accuracy comparison of 7 detectors \emph{w.r.t.} 6 attacks on face datasets MEDS and Multi-PIE, respectively. Here, universal perturbation is imposed on small-scale face images, with 50\%-50\% training-testing split, where both perturbation and content patterns are relatively homogeneous. Even under this protocol, none of the competing methods except TDSC’20 achieved $ > $ 90\% detection accuracy. This implies that non-complete image representations are not sufficient for supporting an efficient detector, even in small-scale detection scenarios. Under such simple experimental protocol, both the proposed detector and TDSC’20 exhibit $\sim$ 100\% accuracy, in line with general expectations of this paper.

\begin{figure}[!t]
	\centering
	\includegraphics[scale=0.48]{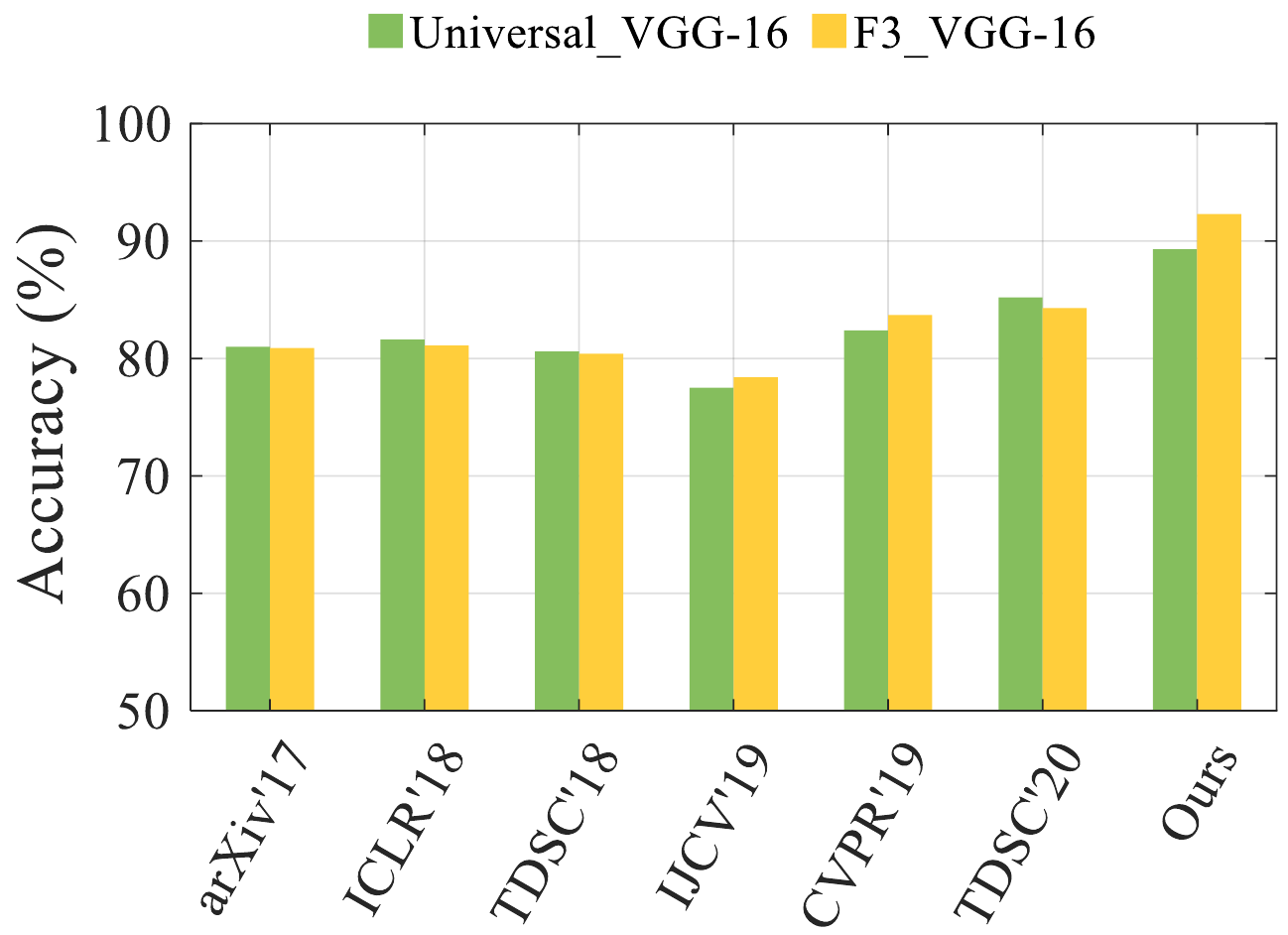}
	\centering
	\caption{The benchmarking of adversarial perturbation detection accuracy for different detectors on the ImageNet dataset.}
\end{figure}

\begin{table}[!t]
	\renewcommand{\arraystretch}{1}
	\centering
	\caption{Recall, Precision, F1, and Accuracy Scores (\%) of Benchmarking Experiments on the ImageNet dataset with More Comprehensive Models and Attacks.}
	\begin{tabular}{cccccc}
		\toprule
		Model & Attack & Recall & Precision & F1 & Accuracy \\
		\midrule
		{\multirow{3}[1]{*}{VGG-16}} & FGSM & 93.59 & 88.89 & 91.18 & 90.95 \\
		~ & BIM & 87.42 & 78.41 & 82.67 & 81.67  \\
		~ & PGD & 82.70 & 80.27 & 81.47 & 81.19 \\ 
		\midrule
		{\multirow{3}[1]{*}{GoogLeNet}} & FGSM & 95.65 & 90.95 & 93.24 & 93.07  \\
		~ & BIM & 90.73 & 80.27 & 85.18 & 84.21  \\
		~ & PGD & 90.97 & 79.92 & 85.08 & 84.05  \\
		\midrule
		{\multirow{3}[1]{*}{ResNet}} & FGSM & 99.11 & 96.66 & 97.87 & 97.84  \\
		~ & BIM & 93.11 & 87.53 & 90.23 & 89.92  \\
		~ & PGD & 97.02 & 93.95 & 95.46 & 95.38  \\
		\midrule
		{\multirow{3}[1]{*}{ViT-B/16}} & FGSM & 96.41 & 95.07 & 95.74 & 95.71  \\
		~ & BIM & 86.33 & 79.80 & 82.94 & 82.24  \\
		~ & PGD & 92.86 & 89.75 & 91.28 & 91.13  \\
		\midrule
		{\multirow{3}[1]{*}{Swin-B}} & FGSM & 99.23 & 98.84 & 99.03 & 99.03  \\
		~ & BIM & 96.09 & 94.34 & 95.21 & 95.16  \\
		~ & PGD & 98.07 & 96.62 & 97.34 & 97.32  \\ 
		\bottomrule
	\end{tabular}
\end{table}

\subsubsection{ImageNet}\label{exp-bench-imgnet}
As for Fig. 9, the accuracy comparison of 7 detectors \emph{w.r.t.} 2 attacks on ImageNet dataset is illustrated. Here, universal perturbation is imposed on large-scale natural images, also of 50\%-50\% training-testing split. Clearly, the diversity of the image content increases significantly \emph{w.r.t.} the previous experimental protocol, increasing also the difficulty of the detection. An interesting phenomenon is that the gap between TDSC’20 and other competing methods becomes smaller. This implies that the simple decision-level fusion of global and local orthogonal transforms in TDSC’20 cannot handle more complex detection tasks well. It should be considered as an inflexible remedy for the contradiction of spatial and frequency discriminability. In general, our method still achieves $\sim$ 5\% gain \emph{w.r.t.} TDSC’20 and $\sim$ 10\% gain \emph{w.r.t.} other competing methods. In Table I, more comprehensive benchmarking scores on the ImageNet dataset are listed. Here, 3 representative attacks are  imposed on large-scale natural images, with 5 models covering recent transformer architectures, also of 50\%-50\% training-testing split. In the total of 15 attack-model pairs, our detector exhibits $ > $ 90\% scores of F1 and accuracy for 9 pairs. Note that all competing detectors in Fig. 9 fail to achieve $ > $ 90\% scores even with the universal perturbation.  For the worst case, the F1 and accuracy scores of our detector are still $\sim$ 80\%, reaching the level of scores for competing detectors in Fig. 9 (but with significantly simpler protocol). It is an important illustration on the strength of our spatial-frequency discriminative detector \emph{w.r.t.} variable perturbations in the presence of complex content interference. Note that we complement more comprehensive results on ImageNet with different intensities in Appendix B.

The above consistent performance gains in Figs. 5 $\sim$ 9 and Table I validate the advanced nature of the proposed detector, revealing the potential of our spatial-frequency discriminative decomposition in perturbation detection tasks. In addition to the above prediction scores, we present the prediction times in Appendix B, which also exhibit real-time potential.

\subsection{Crossing Experiments}\label{exp-cross}

Through above benchmark experiments, we have positioned our detector \emph{w.r.t.} some state-of-the-art detectors. Such extensive results indicate a consistent performance advantage of the proposed method under the typical experiment protocol. In this part, we will further analyze whether the above performance advantage arises from a certain over-fitting. More specifically, we will test a trained detector by crossing to other similar experiment protocols, thereby quantifying its transferability to reasonable changes.

\subsubsection{Crossing Dataset}\label{exp-cross-dataset}
Table II lists the various performance scores of the proposed detector on crossing dataset protocol with universal perturbation. Here, the protocol involves 4 datasets: ImageNet, MEDS, Multi-PIE, and PubFig. In general, after sufficient learning, a promising detector is expected to be generalizable to natural differences of the training and testing phases. As can be observed, for all cases with sufficient learning, our detector exhibits $\sim$ 100\% scores of recall, precision, F1, and accuracy. Here, the detector trained on the very diverse ImageNet can be smoothly transferred to the lower diversity face datasets MEDS, Multi-PIE, and PubFig. Crossing experiments between the 3 face datasets (with similar diversity) also exhibited consistently good scores. Such numerical evidence suggests that our spatial-frequency discriminative decomposition provides complete and intrinsic features of adversarial perturbations, and therefore generalizes well to unseen-but-similar datasets. As can be expected, when the diversity of the training dataset is much smaller than that of the testing one, our detector shows a significant performance degradation. Therefore, such overfitting caused by biased training should be strongly avoided in practical defense, by training with sufficient diversity data.

\begin{table}[!t]
	\renewcommand{\arraystretch}{1}
	\centering
	\caption{ Recall, Precision, F1, and Accuracy Scores (\%) of Crossing Dataset Experiments.}
	\begin{tabular}{cccccc}
		\toprule
		Training & Testing & Recall & Precision & F1 & Accuracy \\
		\midrule
		{\multirow{2}[1]{*}{ImageNet}} & MEDS & 98.60 & 99.86 & 99.22 & 99.23 \\ 
		~ & Multi-PIE & 99.80 & 99.87 & 99.84 & 99.84 \\
		~ & PubFig & 99.67 & 99.80 & 99.74 & 99.74 \\ 
		\midrule
		{\multirow{2}[1]{*}{MEDS}} & ImageNet & 95.04 & 72.94 & 82.54 & 79.90 \\
		~ & Multi-PIE & 100.00 & 100.00 & 100.00 & 100.00 \\
		~ & PubFig & 99.91 & 99.86 & 99.88 & 99.88 \\ 
		\midrule
		{\multirow{2}[1]{*}{Multi-PIE}} & ImageNet & 99.88 & 54.84 & 70.80 & 58.82 \\
		~ & MEDS & 100.00 & 91.52 & 95.57 & 95.37 \\
		~ & PubFig & 99.99 & 99.27 & 99.63 & 99.63 \\ 
		\midrule
		{\multirow{2}[1]{*}{PubFig}} & ImageNet & 98.92 & 68.88 & 81.21 & 77.19 \\
		~ & MEDS & 99.86 & 99.86 & 99.86 & 99.86 \\
		~ & Multi-PIE & 100.00 & 100.00 & 100.00 & 100.00 \\ 
		\bottomrule
	\end{tabular}
\end{table}

\begin{table}[!t]
	\renewcommand{\arraystretch}{1}
	\centering
	\caption{Recall, Precision, F1, and Accuracy Scores (\%) of Crossing Model Experiments.}
	\resizebox{\columnwidth}{!}{
		\begin{tabular}{ccccccc}
			\toprule		
			Training & Testing & Dataset & Recall & Precision & F1 & Accuracy \\ 
			\midrule
			{\multirow{3}[1]{*}{VGG-16}} & {\multirow{3}[1]{*}{GoogleNet}} & MEDS & 100.00 & 97.00 & 98.48 & 98.46 \\
			~ & ~ & Multi-PIE & 100.00 & 99.74 & 99.87 & 99.87 \\
			~ & ~ & PubFig & 99.47 & 99.83 & 99.65 & 99.65 \\ 
			\midrule
			{\multirow{3}[1]{*}{VGG-16}} & {\multirow{3}[1]{*}{CaffeNet}} & MEDS & 90.45 & 96.99 & 93.61 & 93.82 \\ 
			~ & ~ & Multi-PIE & 100.00 & 99.74 & 99.87 & 99.87 \\
			~ & ~ & PubFig & 98.78 & 99.83 & 99.30 & 99.30 \\
			\bottomrule
		\end{tabular}
	}
\end{table}

\begin{table}[!t]
	\renewcommand{\arraystretch}{1}
	\centering
	\caption{ Recall, Precision, F1, and Accuracy Scores (\%) of Crossing Attack Experiments.}
	\begin{tabular}{ccccccc}
		\toprule		
		Training & Testing & Recall & Precision & F1 & Accuracy \\ 
		\midrule
		{\multirow{5}[1]{*}{BIM}} & FGSM & 100.00 & 100.00 & 100.00 & 100.00 \\
		~ & PGD & 100.00 & 99.98 & 99.99 & 99.99 \\
		~ & FAB & 84.82 & 99.53 & 91.59 & 92.21 \\
		~ & Square & 78.82 & 99.50 & 87.96 & 89.21 \\ 
		~ & Jitter & 99.68 & 99.98 & 99.83 & 99.83  \\ 
		\midrule
		{\multirow{5}[1]{*}{FGSM}} & BIM & 99.98 & 99.86 & 99.92 & 99.92 \\
		~ & PGD & 100.00 & 99.86 & 99.93 & 99.93 \\ 
		~ & FAB & 76.90 & 99.82 & 86.87 & 88.38 \\ 
		~ & Square & 77.48 & 99.82 & 87.24 & 88.67 \\ 
		~ & Jitter & 98.46 & 100.00 & 99.22 & 99.23 \\ 
		\midrule
		{\multirow{5}[1]{*}{PGD}} & BIM & 100.00 & 99.86 & 99.93 & 99.93 \\
		~ & FGSM & 100.00 & 99.86 & 99.93 & 99.93 \\
		~ & FAB & 80.28 & 99.83 & 88.99 & 90.07 \\
		~ & Square & 78.70 & 99.82 & 88.01 & 89.28 \\  
		~ & Jitter & 99.78 & 99.86 & 99.82 & 99.82 \\ 
		\midrule
		{\multirow{5}[1]{*}{FAB}} & BIM & 100.00 & 99.11 & 99.55 & 99.55 \\
		~ & FGSM & 100.00 & 99.11 & 99.55 & 99.55 \\
		~ & PGD & 100.00 & 99.11 & 99.55 & 99.55 \\
		~ & Square & 79.06 & 98.87 & 87.86 & 89.08 \\ 
		~ & Jitter & 100.00 & 98.93 & 99.46 & 99.46 \\ 
		\midrule
		{\multirow{5}[1]{*}{Square}} & BIM & 100.00 & 98.68 & 99.33 & 99.33 \\ 
		~ & FGSM & 100.00 & 98.78 & 99.38 & 99.38 \\ 
		~ & PGD & 100.00 & 98.48 & 99.24 & 99.23 \\
		~ & FAB & 89.14 & 98.52 & 93.60 & 93.90 \\ 
		~ & Jitter & 100.00 & 99.44 & 99.72 & 99.72 \\ 
		\midrule
		{\multirow{5}[1]{*}{Jitter}}  & BIM & 100.00 & 99.92 & 99.96 & 99.96  \\
		~ & FGSM & 100.00 & 99.92 & 99.96 & 99.96  \\ 
		~ & PGD & 100.00 & 99.92 & 99.96 & 99.96  \\
		~ & FAB & 86.20 & 99.91 & 92.55 & 93.06  \\ 
		~ & Square & 78.24 & 99.72 & 87.68 & 89.01  \\
		\bottomrule
	\end{tabular}
\end{table}

\emph{Remark.} In addition, we would like to provide some supplementary information as comparison baselines. Based on scores reported in the literature, competing algorithms arXiv’17 and TDSC’18 achieve $\sim$ 80\% accuracy scores on the crossing dataset protocol from MEDS to Multi-PIE, and $\sim$ 70\% accuracy scores for the reverse protocol. Thus they exhibit a $ > $ 20\% performance gap \emph{w.r.t.} the proposed detector. It is further verified that the proposed decomposition serves as a more generic representation than such simple filters.

\subsubsection{Crossing Model}\label{exp-cross-mod}
In Table III, we provide the various performance scores of the proposed detector on crossing model protocol with universal perturbation. Here, our detector is trained on VGG-16 and tested on GoogleNet or CaffeNet, where the training and testing are also considered on three face datasets. Note that although the adversarial perturbations on different models differ significantly at the numerical level, they still exhibit specific statistical consistency; the mining of this consistency largely reflects the generalizability of the detector. Obviously, the proposed detector remains stable, regardless of the different datasets and the different performance metrics. For most cases, our detector exhibits $\sim$ 100\% scores, and even the worst case (VGG-16 to CaffeNet on MEDS) is still 93.82\%. These consistent phenomena suggest that the proposed detector and its foundational decomposition have well generalizability to unseen-but-similar models.

\emph{Remark.} In addition, we would like to provide some supplementary information as comparison baselines. Based on scores reported in the literature, competing algorithm TDSC’20 achieves $\sim$ 93\% accuracy score on the crossing model protocol from VGG-16 to GoogleNet on MEDS, and $\sim$ 96\% accuracy score for the similar protocol on Multi-PIE. Therefore, our detector still exhibits gains \emph{w.r.t.} this orthogonal transform based detector.

\subsubsection{Crossing Attack}\label{exp-cross-att}
As for Table IV, we list the various performance scores of the proposed detector on crossing attack protocol on MNIST dataset. Here, the protocol involves 6 attacks, \emph{i.e.}, BIM, FGSM, PGD, FAB, Square, and Jitter, with quite significant differences in their designs. The experiment will consider any crossing pair of these 6 attacks in training and testing, for a total of 30 pairs. As can be expected, the adversarial perturbations derived from these attacks are different numerically, but meanwhile have similar statistical properties. In the table, the extensive performance scores indicate that the proposed detector is capable of representing the statistical consistency over such attacks. For 20 crossing pairs, our detector exhibits $\sim$ 100\% scores of F1 and accuracy. For the worst case, the F1 and accuracy scores of our detector are still $\sim$ 87\%. In general, the performance of the proposed detector is quite promising, even \emph{w.r.t.} the generalizability of the unseen and somewhat different adversarial perturbations. Note that the detector trained on more complex attacks (\emph{i.e.}, FAB, Square, and Jitter) usually has better transferability for simple attacks (\emph{i.e.}, BIM, FGSM, and PGD). The above common phenomena illustrate the importance of introducing sufficiently diverse data during training. Therefore in the next challenging experiments we will consider the sufficient training with a mixture of multiple attacks to achieve real-world defense.

\emph{Remark.} In addition, we would like to provide some supplementary information as comparison baselines. Based on scores reported in the literature, competing algorithms arXiv’17, ICLR’18, TDSC’18, CVPR’19, and TDSC’20 achieve $\sim$ 64\%, $\sim$ 76\%, $\sim$ 65\%, $\sim$ 68\%, and $\sim$ 95\% average accuracy scores over FGSM-like crossing pairs, respectively. In Table IV, our algorithm exhibits an average accuracy score of $\sim$ 96\%. Note that even though our protocol involves more diverse attacks, the proposed detector still provides better scores than above competing methods. 

The above consistent performance gains in Tables II $\sim$ IV validate the advanced nature of the proposed detector, revealing the generalizability of our spatial-frequency discriminative decomposition for unseen-but-similar scenarios.

\begin{figure*}[!t]
	\centering
	\subfigure[$N=5$]{\includegraphics[scale=0.5]{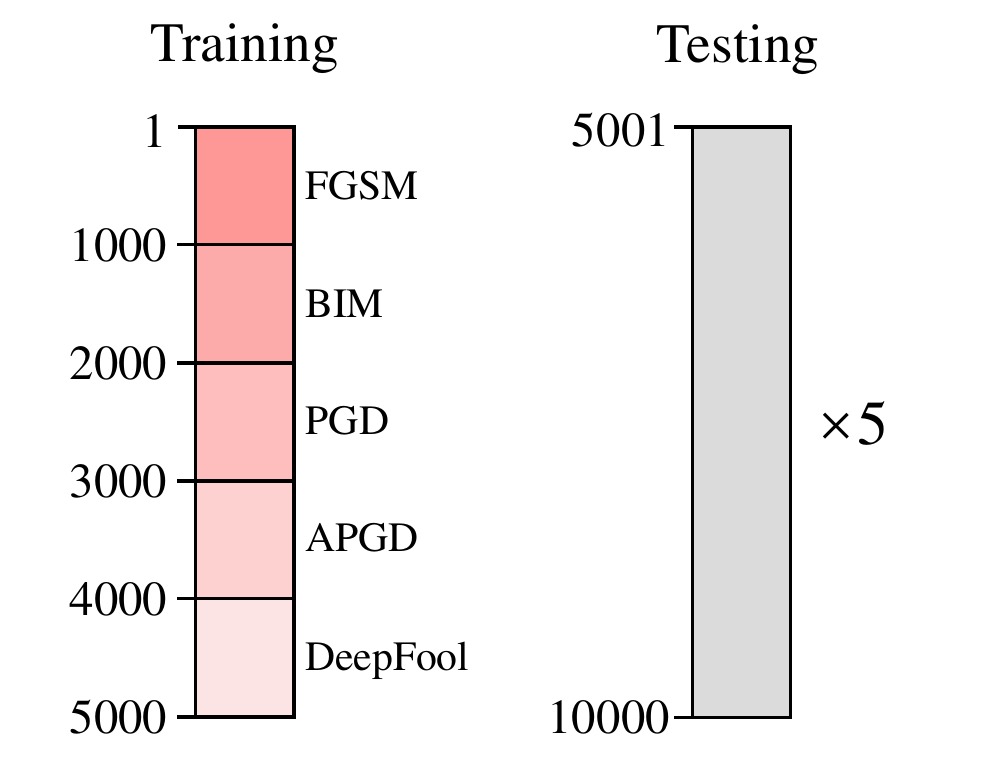}}\hspace{5mm}
	\subfigure[$N=10$]{\includegraphics[scale=0.5]{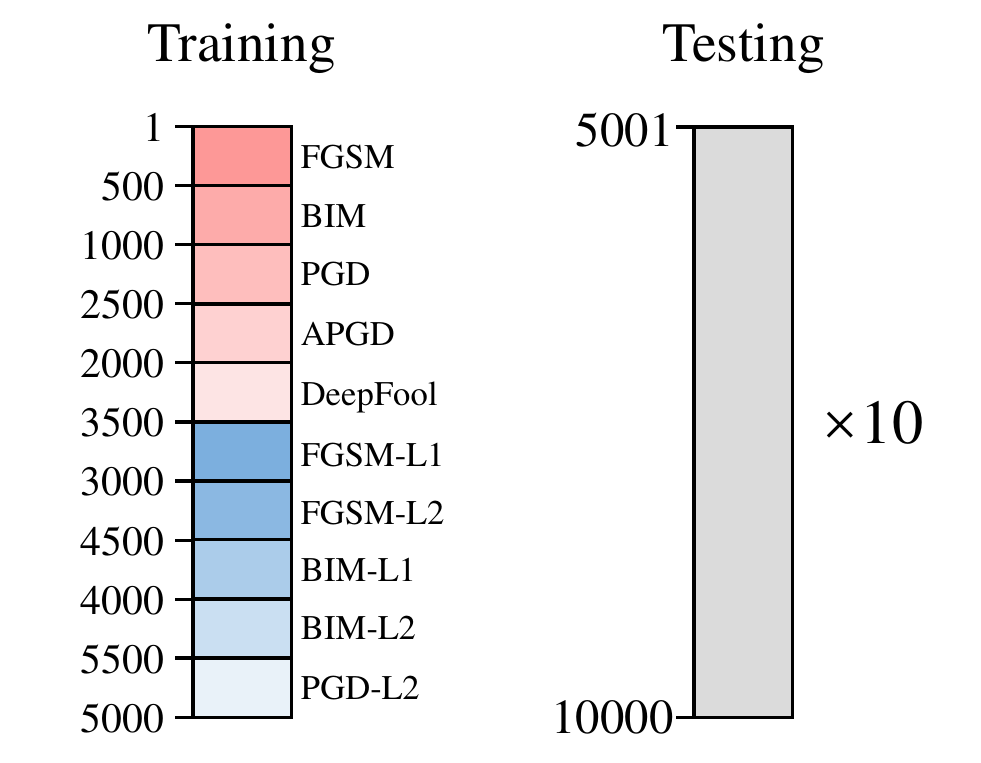}}\hspace{5mm}
	\subfigure[$N=15$]{\includegraphics[scale=0.5]{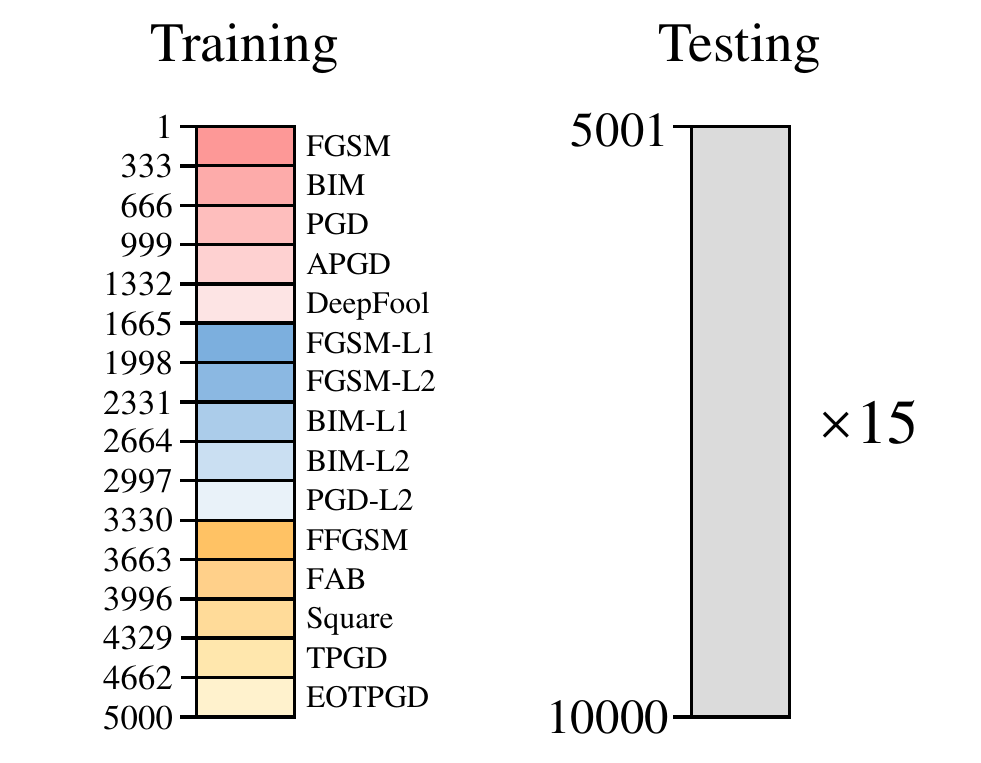}}
	\caption{Illustration for the adversarial image division in challenging experiments. Note that the more attacks involved, the fewer training examples for each attack, and therefore this protocol is a comprehensive challenge for the discriminability, generalization, completeness, and efficiency of the detector.}
\end{figure*}

\begin{figure*}[!t]
	\centering
	\subfigure{\includegraphics[scale=0.45]{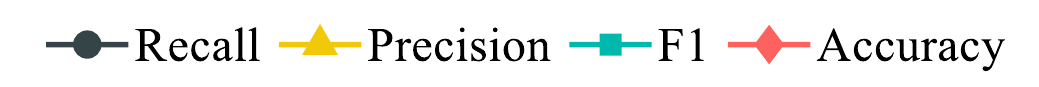}}
	\\
	\subfigure[$N=5$]{\includegraphics[scale=0.43]{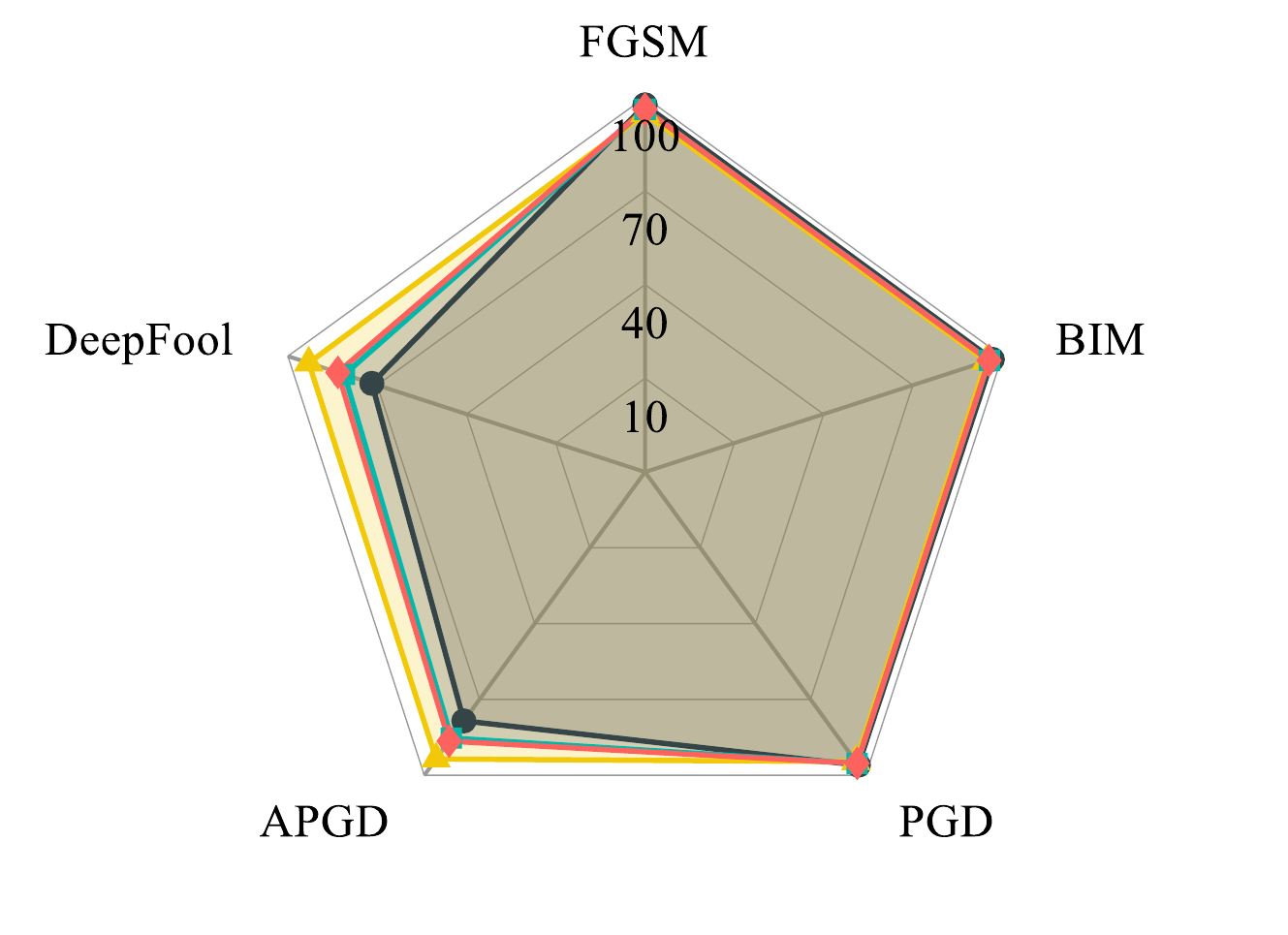}}\hspace{2mm}
	\subfigure[$N=10$]{\includegraphics[scale=0.43]{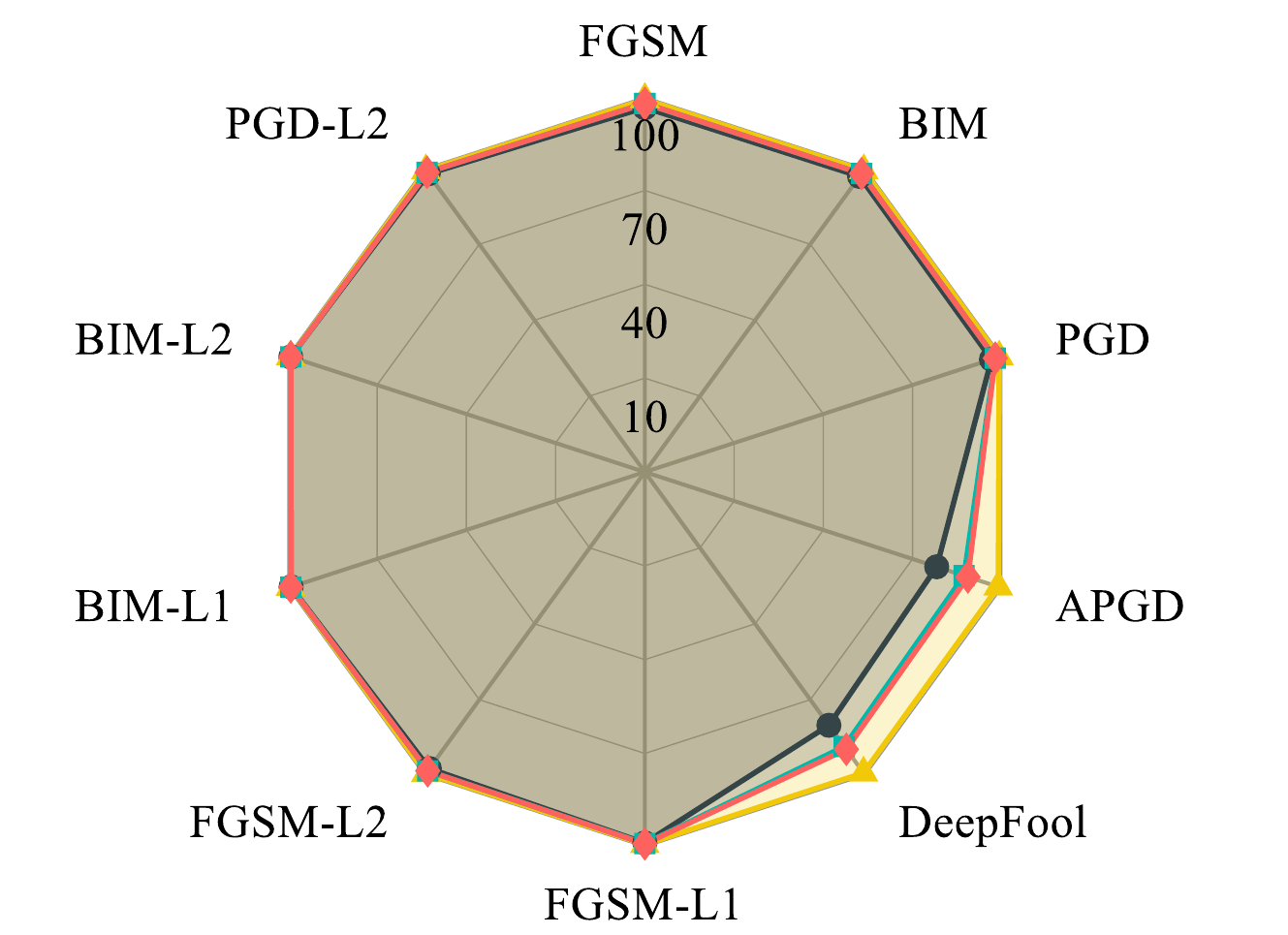}}\hspace{2mm}
	\subfigure[$N=15$]{\includegraphics[scale=0.43]{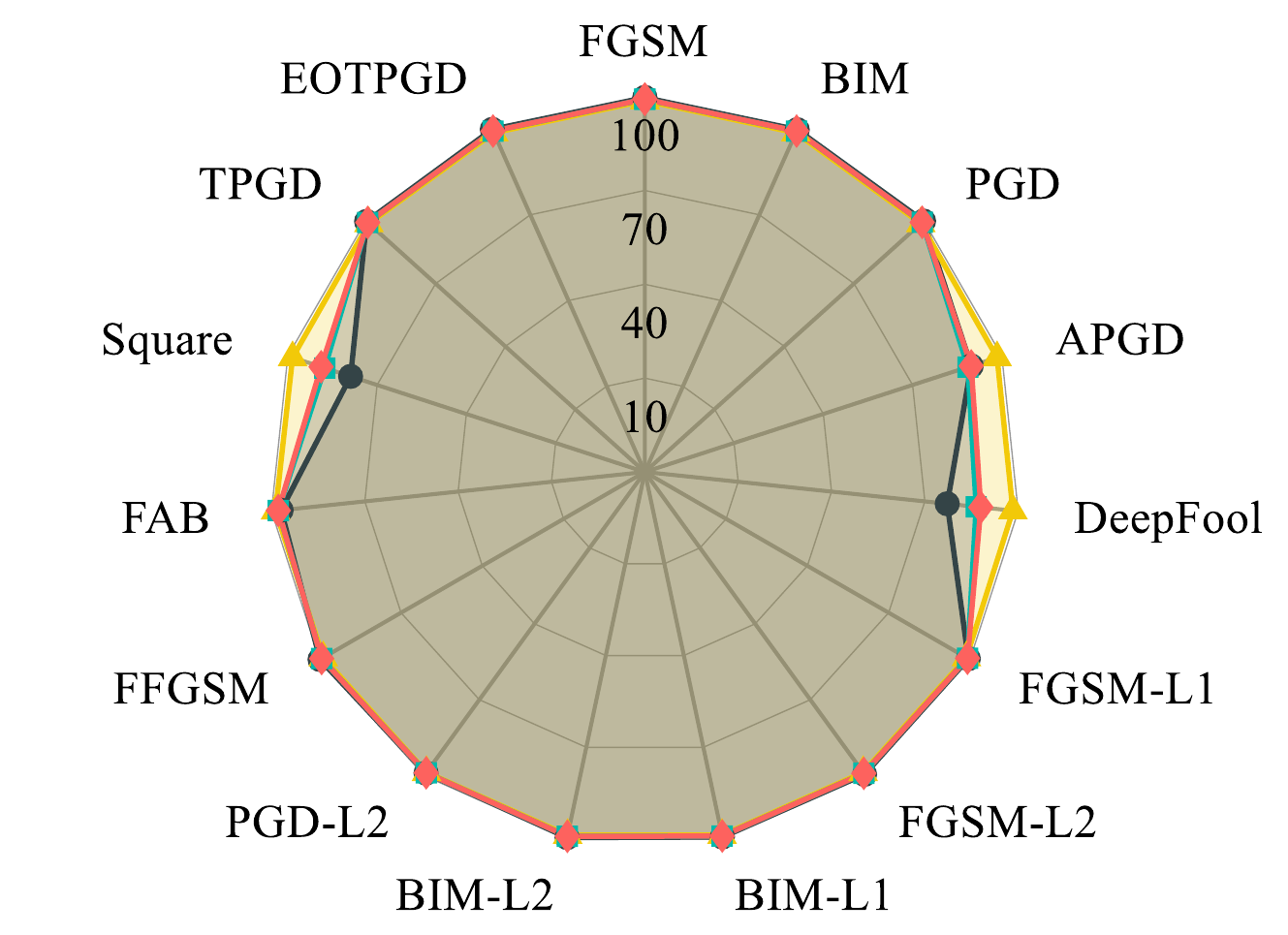}}	
	\centering
	\caption{Recall, precision, F1, and accuracy scores (\%) of challenging experiments. Note that the proposed detector performs consistently even with more attacks and fewer training examples, verifying its discriminability, generalization, completeness, and efficiency.}
\end{figure*}

\subsection{Challenging Experiments}\label{exp-chall}

The above benchmarking and crossing experiments have extensively demonstrated the advantages of the proposed detector over state-of-the-art detectors under ideal protocols. In this part, we turn to more challenging protocols. Such protocols can verify the discriminability, generalization, completeness, and efficiency of the proposed detectors in a comprehensive and realistic manner.

\subsubsection{From Ideal to Practice}\label{exp-chall-mot}

We begin with a discussion on the practical deployment of the adversarial perturbation detector and the corresponding training strategy. In real-world scenarios, it is expected that the detector enables an accurate and generic defense for the widest possible range of adversarial attacks. In general, two training and deployment approaches exist to achieve this goal:
\begin{itemize}
	\item \emph{Integration during inference.} Multiple detectors are involved here, each of which is trained on adversarial examples from only one attack. When deployed, the example under analysis is filled into these detectors separately, and the corresponding results are then integrated as the final prediction. Therefore, this approach can be construed as a late integration strategy.
	\item \emph{Integration during training.} Only one detector is involved here, which is trained directly on adversarial examples from the widest possible range of attacks. After training, the detector is deployed directly to predict arbitrary examples. Therefore, this approach can be construed as an early fusion strategy.
\end{itemize}

It is clear that the integration during training is more compact. Compared to the integration during inference, it will consume significantly fewer computational resources in deployment, while not involving tricky integration policy design for the results from multiple detectors. However, most of the benchmarking and crossing experiments in the literature are trained only with single kind of attack, and thus only verifying the effectiveness for the integration during inference. Therefore, we not only consider benchmarking and crossing experiments for the integration during inference (Sections \ref{exp-bench} and \ref{exp-cross}), but will also focus on more challenging experiments for the integration during training (Section \ref{exp-chall}).

\subsubsection{A Comprehensive and Realistic Protocol}\label{exp-chall-res}

For a comprehensive investigation of the proposed detector in the scenario of integration during training, we design the following experimental protocol. We selected 10000 original images in MNIST for generating the experimental images, in which the first/latter 5000 images are used to derive the training/testing examples respectively. Assuming that $N$ attacks are considered, where $N = 5,10,15$ in our experiments. For the training, the first 5000 images are equally divided into $N$ parts for generating the adversarial images corresponding to per attack, and then they are used as training examples together with the first 5000 original images. Our detector will be trained directly on such 10000 images. For the testing, the examples under each attack will consist of the latter 5000 original images and 5000 corresponding adversarial images, resulting in a total of $N \times 10000$ testing examples.

In Fig. 10, we present an illustration for the above adversarial image division \emph{w.r.t.} $N = 5,10,15$. Note that when $N$ increases, the number of adversarial images from each attack in the training set, \emph{i.e.}, $5000/N$, will decrease, while the counterpart in the testing set remains 5000. Obviously, this protocol is a comprehensive challenge for the discriminability, generalization, completeness, and efficiency of the detector.

In Fig. 11, we show the various performance scores of the proposed detector on challenging experiment protocol. Here, corresponding to Fig. 10, the protocol involves increasing number of attacks and decreasing number of training examples from (a) to (c). In general, a promising detector is expected to be stable to realistic variations in the problem complexity or the training scale. As can be observed here, our detector exhibits consistent recall, precision, F1, and accuracy scores on all three scenarios (a) $\sim$ (c). For the most challenging scenario (c), where number of attacks is 15 and number of training adversarial images per attack is only 333, the scores \emph{w.r.t.} most attacks (except for APGD, DeepFool, and Squara) are even $\sim$ 100\%. The worst case is DeepFool with $\sim$ 87\% scores of F1 and accuracy. Such scores are still generally satisfactory, considering the training/testing adversarial images ratio is 300/5000. The above phenomenon suggests that our detector and its foundational decomposition have:
\begin{itemize}
	\item Discriminability for a wide range of attacks;
	\item Generalization for unseen-but-similar examples;
	\item Completeness for potential perturbation patterns; 
	\item Efficiency on training scale.
\end{itemize}

Starting from such properties, the proposed detector can be regarded as a more comprehensive and effective defense for the realistic scenarios with integration during training.

\begin{figure}[!t]
	\centering
	\begin{minipage}[t]{0.18\linewidth}
		{\includegraphics[width=1.6cm,height=1.6cm]{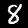}}\vspace{3.pt}
		{\includegraphics[width=1.6cm,height=1.6cm]{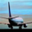}}\vspace{3.pt}
		{\includegraphics[width=1.6cm,height=1.6cm]{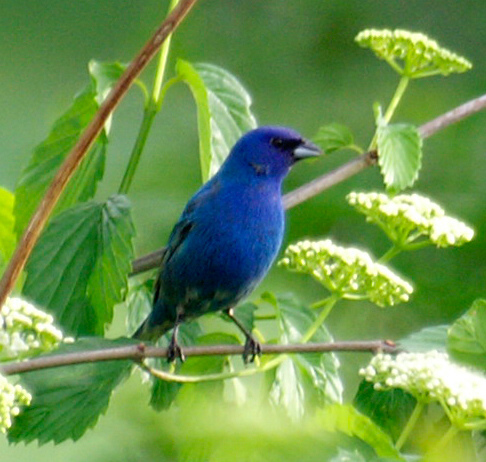}}
		\centerline{\footnotesize Original}
	\end{minipage}
	\begin{minipage}[t]{0.18\linewidth}
		{\includegraphics[width=1.6cm,height=1.6cm]{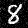}}\vspace{3.pt}
		{\includegraphics[width=1.6cm,height=1.6cm]{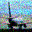}}\vspace{3.pt}
		{\includegraphics[width=1.6cm,height=1.6cm]{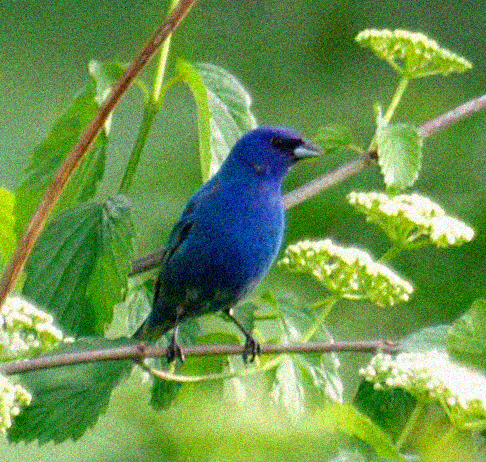}}
		\centerline{\footnotesize Gaussian}
	\end{minipage}
	\begin{minipage}[t]{0.18\linewidth}
		{\includegraphics[width=1.6cm,height=1.6cm]{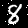}}\vspace{3.pt}
		{\includegraphics[width=1.6cm,height=1.6cm]{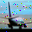}}\vspace{3.pt}
		{\includegraphics[width=1.6cm,height=1.6cm]{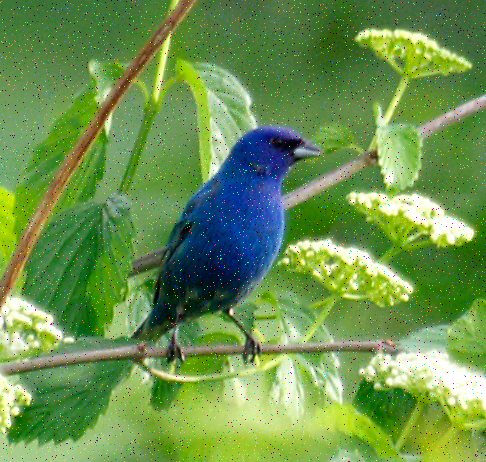}}
		\centerline{\footnotesize S\&P}
	\end{minipage}
	\begin{minipage}[t]{0.18\linewidth}
		{\includegraphics[width=1.6cm,height=1.6cm]{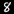}}\vspace{3.pt}
		{\includegraphics[width=1.6cm,height=1.6cm]{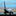}}\vspace{3.pt}
		{\includegraphics[width=1.6cm,height=1.6cm]{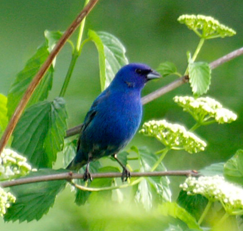}}
		\centerline{\footnotesize Compression}
	\end{minipage}
	\begin{minipage}[t]{0.18\linewidth}
		{\includegraphics[width=1.6cm,height=1.6cm]{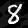}}\vspace{3.pt}
		{\includegraphics[width=1.6cm,height=1.6cm]{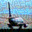}}\vspace{3.pt}
		{\includegraphics[width=1.6cm,height=1.6cm]{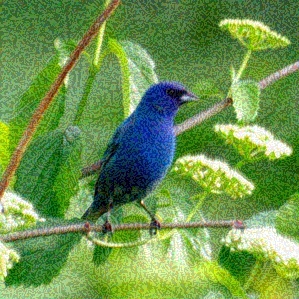}}
		\centerline{\footnotesize Adversarial}
	\end{minipage}
	\caption{Illustration of harmless perturbations (noise and compression) versus adversarial perturbations.}
\end{figure}

\begin{table}[!t]
	\renewcommand{\arraystretch}{1}
	\centering
	\caption{Recall, Precision, F1, and Accuracy Scores (\%) in Distinguishing Between Harmless and Adversarial Perturbations.}
	\begin{tabular}{cccccc}
		\toprule
		 & Dataset & Recall & Precision & F1 & Accuracy \\
		\midrule
		{\multirow{3}[1]{*}{Clean}} & MNIST & 100.00 & 99.98 & 99.99 & 99.99 \\
		~ & CIFAR-10 & 99.24 & 94.17 & 96.64 & 96.55  \\
		~ & ImageNet & 93.59 & 88.89 & 91.18 & 90.95 \\ 
		\midrule
		{\multirow{3}[1]{*}{Gaussian}} & MNIST & 97.96 & 99.94 & 98.94 & 98.95  \\
		~ & CIFAR-10 & 93.96 & 92.88 & 93.42 & 93.38  \\
		~ & ImageNet & 94.07 & 84.77 & 89.18 & 88.59  \\
		\midrule
		{\multirow{3}[1]{*}{S\&P}} & MNIST & 99.86 & 100.00 & 99.93 & 99.93  \\
		~ & CIFAR-10 & 98.86 & 86.08 & 92.03 & 91.44 \\
		~ & ImageNet & 92.10 & 84.53 & 88.15 & 87.62 \\
		\midrule
		{\multirow{3}[1]{*}{Compression}} & MNIST & 100.00 & 93.28 & 96.53 & 96.40  \\
		~ & CIFAR-10 & 99.46 & 98.53 & 98.99 & 98.99  \\
		~ & ImageNet & 96.77 & 91.12 & 93.86 & 93.67  \\
		\bottomrule
	\end{tabular}
\end{table}

\subsection{Limitations and Discussions}\label{exp-limit}

In this part, we provide discussions on the limitations of the defense scope of the proposed detector.

Recalling Conjectures 1 and 2, our detector is better suited for \emph{noise-like} perturbation patterns where the spatial-frequency distribution is far from natural images. Therefore, there may be two limitations on the defense scope: 1) false positives for harmless noise-like perturbations (\emph{e.g.}, noise and compression), and 2) false negatives for adversarial perturbations beyond noise-like patterns.

Regarding harmless noise-like perturbations, false positives can be greatly reduced by introducing harmless perturbed examples into the training phase. As shown in Fig. 12, the 3 types of harmless perturbed examples, \emph{i.e.}, Gaussian noise, salt-and-pepper (S\&P) noise, and compression, share the same training labels as the original clean examples. In table V, we list the various performance scores of the proposed detector with such examples on the 3 datasets. In general, the scores after introducing harmless perturbations do not differ much from the baseline case with only original clean examples. Clearly, our detector is capable of distinguishing between harmless and adversarial perturbations, even with similar noise-like patterns. Therefore, if the defense scenario treats a certain type of perturbation as harmless, it is necessary to introduce its examples in the training of our detector.

Regarding adversarial perturbations beyond noise-like patterns, our detectors do not have good applicability. As demonstrated in the previous experiments, noise-like modeling successfully covers a wide range of adversarial perturbation types, some of which are highly influential mainstream attacks. With the emergence of more diverse attacks, the imperceptibility of perturbations has been greatly expanded conceptually, leading to so-called semantic perturbations (\emph{e.g.}, recoloring as adversarial perturbation \cite{ref89}). Clearly, all methods of checking artifacts at the digital level will fail, covering almost all of the competing detectors listed in this paper. In this regard, future work is open and involves the understanding of semantic artifacts.

\section{Conclusion}\label{conc}

In general, our main goal is to provide a more comprehensive design of adversarial perturbation detector. As a technical foundation of the detector, we have proposed the spatial-frequency discriminative decomposition with secret keys, motivated by the accuracy and security issues of existing detectors. Here, the accuracy and security ingredients in this paper can be summarized as follows.
\begin{itemize}
	\item Regarding the \emph{accuracy}, we attribute the accuracy bottleneck of existing detectors to the fundamental contradiction of spatial and frequency discriminability in the decomposition. Specifically, the non-orthogonal decomposition (\emph{e.g.}, SRM) is not sufficient to completely represent a wide range of potential perturbations. The global (\emph{e.g.}, DST) or local (\emph{e.g.}, DWT) orthogonal decomposition cannot mine both frequency and spatial patterns (Conjectures 1 and 2), thereby failing to fully reveal the statistical regularity (Formulation 2). In this paper, we have introduced the Krawtchouk basis (Definition 2) for more discriminative decomposition, providing a mid-scale representation with rich spatial-frequency information (Property 1). Therefore, the resulting detector exhibits better discriminability in the decomposition of natural images and adversarial perturbations (Main Result 1).
	\item Regarding the \emph{security}, we attribute the defense-aware attack to the transparency of detector-interested features for the attacker. With such knowledge, the attacker can regenerate adversarial perturbations without exhibiting obvious artifacts on these features (Formulations 3 and 4). In this paper, we have proposed the random feature selection for secrecy, where a key controlled pseudorandom number generator determines the spatial and frequency parameters of the decomposition. Therefore, the resulting detector is more secure against the defense-aware attack: it is more difficult to divide between to-be-attacked and to-be-evaded features (Main Result 2).
\end{itemize}

We have provided statistical comparisons with state-of-the-art detectors, by the benchmarking (Section \ref{exp-bench}), crossing (Section \ref{exp-cross}), and challenging (Section \ref{exp-chall}) experiments for both ideal and realistic scenarios (\emph{w.r.t.} integration during inference and training). In general, such extensive experimental results conﬁrm the effectiveness of our detector, exhibiting quite satisfactory discriminability, generalization, completeness, and efficiency \emph{w.r.t.} existing works.

Our future work will focus on more formal statistical analysis for the decomposition of natural images and adversarial perturbations, potentially involving coefficient statistical modeling and hypothesis testing.


\ifCLASSOPTIONcaptionsoff
  \newpage
\fi



%

\bibliographystyle{IEEEtran}
\bibliography{paper}

\begin{thebibliography}{10}
\providecommand{\url}[1]{#1}
\csname url@samestyle\endcsname
\providecommand{\newblock}{\relax}
\providecommand{\bibinfo}[2]{#2}
\providecommand{\BIBentrySTDinterwordspacing}{\spaceskip=0pt\relax}
\providecommand{\BIBentryALTinterwordstretchfactor}{4}
\providecommand{\BIBentryALTinterwordspacing}{\spaceskip=\fontdimen2\font plus
\BIBentryALTinterwordstretchfactor\fontdimen3\font minus
  \fontdimen4\font\relax}
\providecommand{\BIBforeignlanguage}[2]{{%
\expandafter\ifx\csname l@#1\endcsname\relax
\typeout{** WARNING: IEEEtran.bst: No hyphenation pattern has been}%
\typeout{** loaded for the language `#1'. Using the pattern for}%
\typeout{** the default language instead.}%
\else
\language=\csname l@#1\endcsname
\fi
#2}}
\providecommand{\BIBdecl}{\relax}
\BIBdecl

\bibitem{ref1}
Y.~Bengio, A.~Courville, and P.~Vincent, ``Representation learning: A review
  and new perspectives,'' \emph{IEEE Trans. Pattern Anal. Mach. Intell.},
  vol.~35, no.~8, pp. 1798--1828, 2013.

\bibitem{ref2}
L.~Zhao, Q.~Wang, C.~Wang, Q.~Li, C.~Shen, and B.~Feng, ``Veriml: Enabling
  integrity assurances and fair payments for machine learning as a service,''
  \emph{IEEE Trans. Parallel Distrib. Syst.}, vol.~32, no.~10, pp. 2524--2540,
  2021.

\bibitem{ref3}
J.~M. Wing, ``Trustworthy {AI},'' \emph{Commun. ACM}, vol.~64, no.~10, pp.
  64--71, 2021.

\bibitem{ref4}
A.~Fawzi, S.-M. Moosavi-Dezfooli, and P.~Frossard, ``The robustness of deep
  networks: A geometrical perspective,'' \emph{IEEE Signal Process Mag.},
  vol.~34, no.~6, pp. 50--62, 2017.

\bibitem{ref5}
I.~Goodfellow, P.~McDaniel, and N.~Papernot, ``Making machine learning robust
  against adversarial inputs,'' \emph{Commun. ACM}, vol.~61, no.~7, pp. 56--66,
  2018.

\bibitem{ref6}
C.~Szegedy, W.~Zaremba, I.~Sutskever, J.~Bruna, D.~Erhan, I.~Goodfellow, and
  R.~Fergus, ``Intriguing properties of neural networks,'' in \emph{Proc. Int.
  Conf. Learn. Represent.}, 2014.

\bibitem{ref7}
A.~Kurakin, I.~Goodfellow, and S.~Bengio, ``Adversarial examples in the
  physical world,'' in \emph{Artificial intelligence safety and
  security}.\hskip 1em plus 0.5em minus 0.4em\relax Chapman and Hall/CRC, 2018,
  pp. 99--112.

\bibitem{ref8}
J.~Su, D.~V. Vargas, and K.~Sakurai, ``One pixel attack for fooling deep neural
  networks,'' \emph{IEEE Trans. Evol. Comput.}, vol.~23, no.~5, pp. 828--841,
  2019.

\bibitem{ref9}
I.~Goodfellow, J.~Shlens, and C.~Szegedy, ``Explaining and harnessing
  adversarial examples,'' in \emph{Proc. Int. Conf. Learn. Represent.}, 2015.

\bibitem{ref10}
N.~Carlini and D.~Wagner, ``Towards evaluating the robustness of neural
  networks,'' in \emph{Proc. IEEE Symp. Secur. Priv.}, 2017, pp. 39--57.

\bibitem{ref11}
S.-M. Moosavi-Dezfooli, A.~Fawzi, O.~Fawzi, and P.~Frossard, ``Universal
  adversarial perturbations,'' in \emph{Proc. IEEE Conf. Comput. Vis. Pattern
  Recognit.}, 2017, pp. 1765--1773.

\bibitem{ref12}
P.-Y. Chen, H.~Zhang, Y.~Sharma, J.~Yi, and C.-J. Hsieh, ``Zoo: Zeroth order
  optimization based black-box attacks to deep neural networks without training
  substitute models,'' in \emph{Proc. ACM Workshop Artif. Intell. Secur.},
  2017, pp. 15--26.

\bibitem{ref13}
K.~Eykholt, I.~Evtimov, E.~Fernandes, B.~Li, A.~Rahmati, C.~Xiao, A.~Prakash,
  T.~Kohno, and D.~Song, ``Robust physical-world attacks on deep learning
  visual classification,'' in \emph{Proc. IEEE Conf. Comput. Vis. Pattern
  Recognit.}, 2018, pp. 1625--1634.

\bibitem{ref79}
P.~Li, Y.~Zhang, L.~Yuan, J.~Zhao, X.~Xu, and X.~Zhang, ``Adversarial attacks
  on video object segmentation with hard region discovery,'' \emph{IEEE Trans.
  Circuits Syst. Video Technol.}, 2023.

\bibitem{ref80}
Z.~Zhou, Y.~Sun, Q.~Sun, C.~Li, and Z.~Ren, ``Only once attack: Fooling the
  tracker with adversarial template,'' \emph{IEEE Trans. Circuits Syst. Video
  Technol.}, 2023.

\bibitem{ref81}
T.~Chen and Z.~Ma, ``Towards robust neural image compression: Adversarial
  attack and model finetuning,'' \emph{IEEE Transactions on Circuits and
  Systems for Video Technology}, 2023.

\bibitem{ref82}
X.~Qin, B.~Li, S.~Tan, W.~Tang, and J.~Huang, ``Gradually enhanced adversarial
  perturbations on color pixel vectors for image steganography,'' \emph{IEEE
  Trans. Circuits Syst. Video Technol.}, vol.~32, no.~8, pp. 5110--5123, 2022.

\bibitem{ref83}
B.~Wang, M.~Zhao, W.~Wang, X.~Dai, Y.~Li, and Y.~Guo, ``Adversarial analysis
  for source camera identification,'' \emph{IEEE Trans. Circuits Syst. Video
  Technol.}, vol.~31, no.~11, pp. 4174--4186, 2020.

\bibitem{ref84}
J.~Zhang, J.~Wang, H.~Wang, and X.~Luo, ``Self-recoverable adversarial
  examples: a new effective protection mechanism in social networks,''
  \emph{IEEE Trans. Circuits Syst. Video Technol.}, vol.~33, no.~2, pp.
  562--574, 2022.

\bibitem{ref14}
M.~Andriushchenko and N.~Flammarion, ``Understanding and improving fast
  adversarial training,'' \emph{Proc. Conf. Neural Inf. Proces. Syst.},
  vol.~33, pp. 16\,048--16\,059, 2020.

\bibitem{ref15}
H.~Liu, Y.~Wang, W.~Fan, X.~Liu, Y.~Li, S.~Jain, Y.~Liu, A.~Jain, and J.~Tang,
  ``Trustworthy {AI}: A computational perspective,'' \emph{ACM Trans. Intell.
  Syst. Technolog.}, vol.~14, no.~1, pp. 1--59, 2022.

\bibitem{ref16}
C.~Xiang, A.~N. Bhagoji, V.~Sehwag, and P.~Mittal, ``Patch{G}uard: A provably
  robust defense against adversarial patches via small receptive fields and
  masking.'' in \emph{Proc. USENIX Secur. Symp.}, 2021, pp. 2237--2254.

\bibitem{ref17}
A.~Nayebi and S.~Ganguli, ``Biologically inspired protection of deep networks
  from adversarial attacks,'' \emph{arXiv preprint arXiv:1703.09202}, 2017.

\bibitem{ref18}
A.~Ross and F.~Doshi-Velez, ``Improving the adversarial robustness and
  interpretability of deep neural networks by regularizing their input
  gradients,'' in \emph{Proc. AAAI Conf. Artif. Intell.}, vol.~32, no.~1, 2018.

\bibitem{ref72}
M.~Lecuyer, V.~Atlidakis, R.~Geambasu, D.~Hsu, and S.~Jana, ``Certified
  robustness to adversarial examples with differential privacy,'' in
  \emph{Proc. IEEE Symp. Secur. Priv.}, 2019, pp. 656--672.

\bibitem{ref73}
J.~Cohen, E.~Rosenfeld, and Z.~Kolter, ``Certified adversarial robustness via
  randomized smoothing,'' in \emph{Proc. Int. Conf. Mach. Learn.}, 2019, pp.
  1310--1320.

\bibitem{ref74}
A.~Cullen, P.~Montague, S.~Liu, S.~Erfani, and B.~Rubinstein, ``Double bubble,
  toil and trouble: enhancing certified robustness through transitivity,''
  \emph{Proc. Adv. Neural Inf. Proces. Syst.}, vol.~35, pp. 19\,099--19\,112,
  2022.

\bibitem{ref19}
H.~Zhang, Y.~Yu, J.~Jiao, E.~Xing, L.~El~Ghaoui, and M.~Jordan, ``Theoretically
  principled trade-off between robustness and accuracy,'' in \emph{Proc. Int.
  Conf. Mach. Learn.}, 2019, pp. 7472--7482.

\bibitem{ref20}
X.~Zhang, X.~Zheng, and W.~Mao, ``Adversarial perturbation defense on deep
  neural networks,'' \emph{ACM Comput. Surv.}, vol.~54, no.~8, pp. 1--36, 2021.

\bibitem{ref21}
S.~Qi, Y.~Zhang, C.~Wang, J.~Zhou, and X.~Cao, ``A principled design of image
  representation: Towards forensic tasks,'' \emph{IEEE Trans. Pattern Anal.
  Mach. Intell.}, 2022.

\bibitem{ref22}
G.-L. Chen and C.-C. Hsu, ``Jointly defending {D}eep{F}ake manipulation and
  adversarial attack using decoy mechanism,'' \emph{IEEE Trans. Pattern Anal.
  Mach. Intell.}, 2023.

\bibitem{ref23}
B.~Liang, H.~Li, M.~Su, X.~Li, W.~Shi, and X.~Wang, ``Detecting adversarial
  image examples in deep neural networks with adaptive noise reduction,''
  \emph{IEEE Trans. Dependable Secure Comput.}, vol.~18, no.~1, pp. 72--85,
  2018.

\bibitem{ref24}
J.~Liu, W.~Zhang, Y.~Zhang, D.~Hou, Y.~Liu, H.~Zha, and N.~Yu, ``Detection
  based defense against adversarial examples from the steganalysis point of
  view,'' in \emph{Proc. IEEE Conf. Comput. Vis. Pattern Recognit.}, 2019, pp.
  4825--4834.

\bibitem{ref78}
C.~Guo, J.~S. Frank, and K.~Q. Weinberger, ``Low frequency adversarial
  perturbation,'' in \emph{Proc. Uncertainty Artif. Intell.}, 2020, pp.
  1127--1137.

\bibitem{ref25}
A.~N. Bhagoji, D.~Cullina, C.~Sitawarin, and P.~Mittal, ``Enhancing robustness
  of machine learning systems via data transformations,'' in \emph{Proc. Annu.
  Conf. Inf. Sci. Syst.}, 2018, pp. 1--5.

\bibitem{ref26}
N.~Akhtar, J.~Liu, and A.~Mian, ``Defense against universal adversarial
  perturbations,'' in \emph{Proc. IEEE Conf. Comput. Vis. Pattern Recognit.},
  2018, pp. 3389--3398.

\bibitem{ref27}
A.~Agarwal, R.~Singh, M.~Vatsa, and N.~Ratha, ``Image transformation-based
  defense against adversarial perturbation on deep learning models,''
  \emph{IEEE Trans. Dependable Secure Comput.}, vol.~18, no.~5, pp. 2106--2121,
  2020.

\bibitem{ref28}
A.~Agarwal, G.~Goswami, M.~Vatsa, R.~Singh, and N.~K. Ratha, ``Damad: Database,
  attack, and model agnostic adversarial perturbation detector,'' \emph{IEEE
  Trans. Neural Networks Learn. Syst.}, vol.~33, no.~8, pp. 3277--3289, 2021.

\bibitem{ref30}
S.~G. Mallat, ``A theory for multiresolution signal decomposition: the wavelet
  representation,'' \emph{IEEE Trans. Pattern Anal. Mach. Intell.}, vol.~11,
  no.~7, pp. 674--693, 1989.

\bibitem{ref31}
Z.~Chen, B.~Tondi, X.~Li, R.~Ni, Y.~Zhao, and M.~Barni, ``Secure detection of
  image manipulation by means of random feature selection,'' \emph{IEEE Trans.
  Inf. Forensics Secur.}, vol.~14, no.~9, pp. 2454--2469, 2019.

\bibitem{ref45}
Z.~Zha, X.~Yuan, J.~Zhou, C.~Zhu, and B.~Wen, ``Image restoration via
  simultaneous nonlocal self-similarity priors,'' \emph{IEEE Trans. Image
  Process.}, vol.~29, pp. 8561--8576, 2020.

\bibitem{ref76}
F.~Wu, W.~Yang, L.~Xiao, and J.~Zhu, ``Adaptive wiener filter and natural noise
  to eliminate adversarial perturbation,'' \emph{Electronics}, vol.~9, no.~10,
  p. 1634, 2020.

\bibitem{ref77}
V.~Veerabadran, J.~Goldman, S.~Shankar, B.~Cheung, N.~Papernot, A.~Kurakin,
  I.~Goodfellow, J.~Shlens, J.~Sohl-Dickstein, M.~C. Mozer \emph{et~al.},
  ``Subtle adversarial image manipulations influence both human and machine
  perception,'' \emph{Nat. Commun.}, vol.~14, no.~1, p. 4933, 2023.

\bibitem{ref75}
E.~P. Simoncelli and B.~A. Olshausen, ``Natural image statistics and neural
  representation,'' \emph{Annu. Rev. Neurosci.}, vol.~24, no.~1, pp.
  1193--1216, 2001.

\bibitem{ref46}
S.~Qi, Y.~Zhang, C.~Wang, J.~Zhou, and X.~Cao, ``A survey of orthogonal moments
  for image representation: theory, implementation, and evaluation,'' \emph{ACM
  Comput. Surv.}, vol.~55, no.~1, pp. 1--35, 2021.

\bibitem{ref47}
P.-T. Yap, R.~Paramesran, and S.-H. Ong, ``Image analysis by {K}rawtchouk
  moments,'' \emph{IEEE Trans. Image Process.}, vol.~12, no.~11, pp.
  1367--1377, 2003.

\bibitem{ref48}
H.~Yang, S.~Qi, J.~Tian, P.~Niu, and X.~Wang, ``Robust and discriminative image
  representation: Fractional-order {J}acobi-{F}ourier moments,'' \emph{Pattern
  Recognit.}, vol. 115, p. 107898, 2021.

\bibitem{ref49}
Y.~LeCun, L.~Bottou, Y.~Bengio, and P.~Haffner, ``Gradient-based learning
  applied to document recognition,'' \emph{Proc. IEEE}, vol.~86, no.~11, pp.
  2278--2324, 1998.

\bibitem{ref50}
K.~Simonyan and A.~Zisserman, ``Very deep convolutional networks for
  large-scale image recognition,'' in \emph{Proc. Int. Conf. Learn.
  Represent.}, 2015.

\bibitem{ref51}
C.~Szegedy, W.~Liu, Y.~Jia, P.~Sermanet, S.~Reed, D.~Anguelov, D.~Erhan,
  V.~Vanhoucke, and A.~Rabinovich, ``Going deeper with convolutions,'' in
  \emph{Proc. IEEE Conf. Comput. Vis. Pattern Recognit.}, 2015, pp. 1--9.

\bibitem{ref52}
Y.~Jia, E.~Shelhamer, J.~Donahue, S.~Karayev, J.~Long, R.~Girshick,
  S.~Guadarrama, and T.~Darrell, ``Caffe: Convolutional architecture for fast
  feature embedding,'' in \emph{Proc. ACM Int. Conf. Multimed.}, 2014, pp.
  675--678.

\bibitem{ref85}
K.~He, X.~Zhang, S.~Ren, and J.~Sun, ``Deep residual learning for image
  recognition,'' \emph{Proc. IEEE Conf. Comput. Vis. Pattern Recognit.}, pp.
  770--778, 2016.

\bibitem{ref86}
A.~Dosovitskiy, L.~Beyer, A.~Kolesnikov, D.~Weissenborn, X.~Zhai,
  T.~Unterthiner, M.~Dehghani, M.~Minderer, G.~Heigold, S.~Gelly \emph{et~al.},
  ``An image is worth 16x16 words: transformers for image recognition at
  scale,'' \emph{Proc. Int. Conf. Learn. Represent.}, 2020.

\bibitem{ref87}
Z.~Liu, Y.~Lin, Y.~Cao, H.~Hu, Y.~Wei, Z.~Zhang, S.~Lin, and B.~Guo, ``Swin
  transformer: hierarchical vision transformer using shifted windows,''
  \emph{Proc. IEEE Int. Conf. Comput. Vis.}, pp. 10\,012--10\,022, 2021.

\bibitem{ref53}
Y.~LeCun, C.~Corinna, and J.~B. Christopher, ``{MNIST},''
  \url{http://yann.lecun.com/exdb/mnist/}.

\bibitem{ref54}
A.~Krizhevsky, V.~Nair, and G.~Hinton, ``{CIFAR-10},''
  \url{https://www.cs.toronto.edu/~kriz/cifar.html}.

\bibitem{ref55}
A.~Founds, N.~Orlans, G.~Whiddon, and C.~Watson, ``{MEDS},'' \url{
  https://www.nist.gov/itl/iad/image-group/special-database-32-multiple-encounter-dataset-meds}.

\bibitem{ref56}
R.~Gross, I.~Matthews, J.~Cohn, T.~Kanade, and S.~Baker, ``{Multi-PIE},''
  \url{https://www.cs.cmu.edu/afs/cs/project/PIE/MultiPie/Multi-Pie/Home.html}.

\bibitem{ref57}
N.~Kumar, A.~C. Berg, P.~N. Belhumeur, and S.~K. Nayar, ``{PubFig},''
  \url{https://www.cs.columbia.edu/CAVE/databases/pubfig/}.

\bibitem{ref58}
J.~Deng, W.~Dong, R.~Socher, L.-J. Li, K.~Li, and L.~Fei-Fei, ``{ImageNet},''
  \url{https://www.image-net.org/index.php}.

\bibitem{ref59}
A.~Madry, A.~Makelov, L.~Schmidt, D.~Tsipras, and A.~Vladu, ``Towards deep
  learning models resistant to adversarial attacks,'' in \emph{Proc. Int. Conf.
  Learn. Represent.}, 2018.

\bibitem{ref60}
F.~Croce and M.~Hein, ``Reliable evaluation of adversarial robustness with an
  ensemble of diverse parameter-free attacks,'' in \emph{Proc. Int. Conf. Mach.
  Learn.}, 2020, pp. 2206--2216.

\bibitem{ref61}
S.-M. Moosavi-Dezfooli, A.~Fawzi, and P.~Frossard, ``Deepfool: a simple and
  accurate method to fool deep neural networks,'' in \emph{Proc. IEEE Conf.
  Comput. Vis. Pattern Recognit.}, 2016, pp. 2574--2582.

\bibitem{ref62}
E.~Wong, L.~Rice, and J.~Z. Kolter, ``Fast is better than free: Revisiting
  adversarial training,'' in \emph{Proc. Int. Conf. Learn. Represent.}, 2020.

\bibitem{ref63}
F.~Croce and M.~Hein, ``Minimally distorted adversarial examples with a fast
  adaptive boundary attack,'' in \emph{Proc. Int. Conf. Mach. Learn.}, 2020,
  pp. 2196--2205.

\bibitem{ref64}
M.~Andriushchenko, F.~Croce, N.~Flammarion, and M.~Hein, ``Square attack: a
  query-efficient black-box adversarial attack via random search,'' in
  \emph{Proc. Eur. Conf. Comput. Vis.}, 2020, pp. 484--501.

\bibitem{ref65}
H.~Zhang, Y.~Yu, J.~Jiao, E.~Xing, L.~El~Ghaoui, and M.~Jordan, ``Theoretically
  principled trade-off between robustness and accuracy,'' in \emph{Proc. Int.
  Conf. Mach. Learn.}, 2019, pp. 7472--7482.

\bibitem{ref66}
R.~S. Zimmermann, ``Comment on "{A}dv-{BNN}: Improved adversarial defense
  through robust {B}ayesian {N}eural {N}etwork",'' \emph{arXiv preprint
  arXiv:1907.00895}, 2019.

\bibitem{ref67}
K.~R. Mopuri, U.~Garg, and R.~V. Babu, ``Fast {F}eature {F}ool: A data
  independent approach to universal adversarial perturbations,'' in \emph{Proc.
  Br. Mach. Vis. Conf.}, 2017.

\bibitem{ref88}
L.~Schwinn, R.~Raab, A.~Nguyen, D.~Zanca, and B.~Eskofier, ``Exploring
  misclassifications of robust neural networks to enhance adversarial
  attacks,'' \emph{Appl. Intell.}, vol.~53, no.~17, pp. 19\,843--19\,859, 2023.

\bibitem{ref68}
R.~Feinman, R.~R. Curtin, S.~Shintre, and A.~B. Gardner, ``Detecting
  adversarial samples from artifacts,'' \emph{arXiv preprint arXiv:1703.00410},
  2017.

\bibitem{ref69}
S.~Liang, Y.~Li, and R.~Srikant, ``Principled detection of out-of-distribution
  examples in neural networks,'' in \emph{Proc. Int. Conf. Learn. Represent.},
  2018.

\bibitem{ref70}
G.~Goswami, A.~Agarwal, N.~Ratha, R.~Singh, and M.~Vatsa, ``Detecting and
  mitigating adversarial perturbations for robust face recognition,''
  \emph{Int. J. Comput. Vision}, vol. 127, pp. 719--742, 2019.

\bibitem{ref71}
S.~Ma, Y.~Liu, G.~Tao, W.-C. Lee, and X.~Zhang, ``Nic: Detecting adversarial
  samples with neural network invariant checking,'' in \emph{Proc. Net.
  Distribut. Sys. Secur. Symp.}, 2019.

\bibitem{ref89}
S.~Yuan, Q.~Zhang, L.~Gao, Y.~Cheng, and J.~Song, ``Natural color fool: Towards
  boosting black-box unrestricted attacks,'' in \emph{Proc. Adv. Neural Inf.
  Proces. Syst.}, vol.~35, 2022, pp. 7546--7560.

\end{thebibliography}



%
%
\begin{IEEEbiography}[{\includegraphics[width=1in,height=1.25in,clip,keepaspectratio]{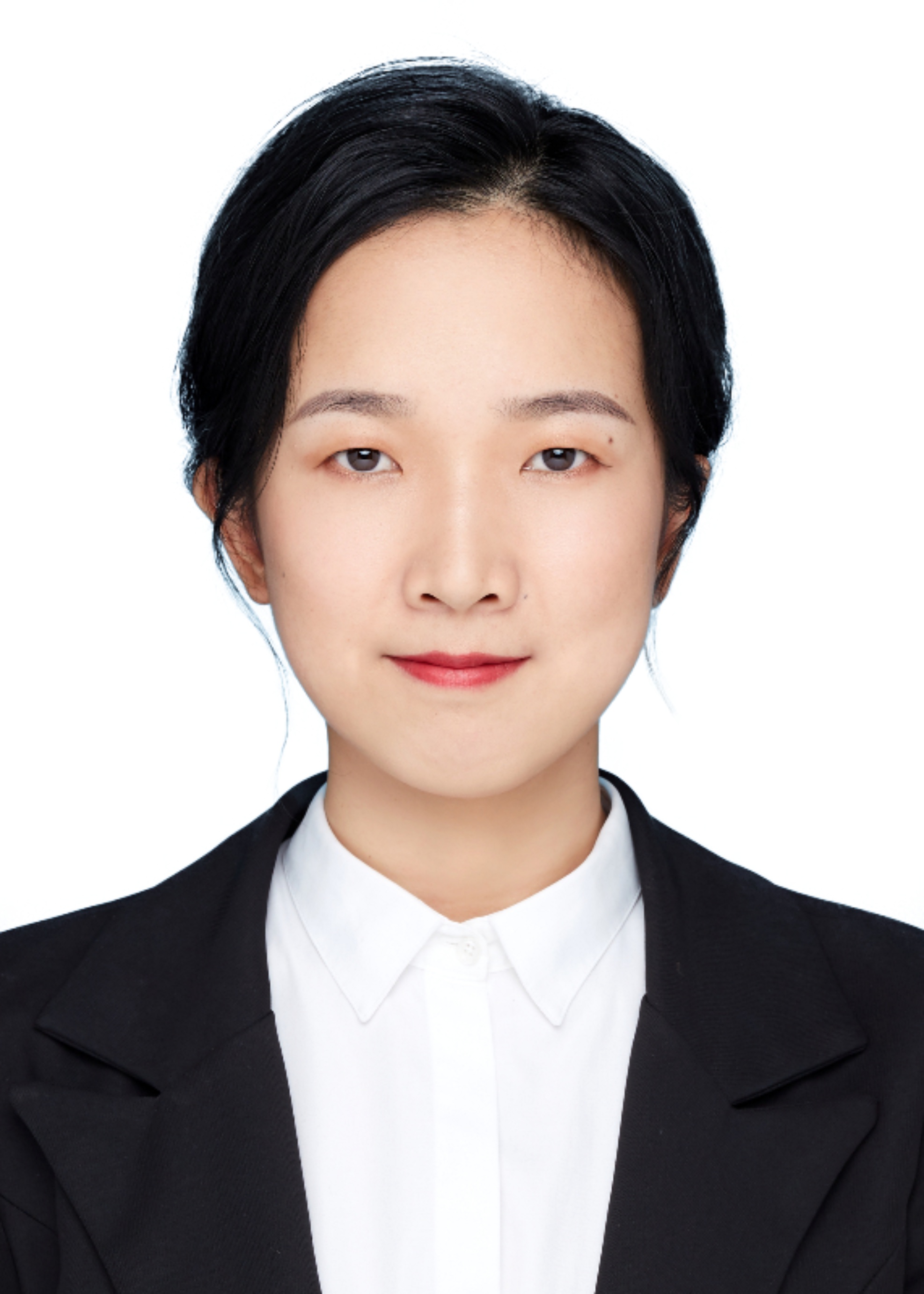}}]{Chao Wang}
	received the B.S. and M.S. degrees from Liaoning Normal University, Dalian, China, in 2017 and 2020, respectively. She is currently pursuing the Ph.D. degree in computer science at Nanjing University of Aeronautics and Astronautics, Nanjing, China. She has authored or coauthored academic papers in top-tier venues including \emph{IEEE Transactions on Information Forensics and Security} and \emph{IEEE Transactions on Pattern Analysis and Machine Intelligence}. Her research interests include trustworthy artificial intelligence, adversarial learning, and media forensics.
\end{IEEEbiography}

\begin{IEEEbiography}[{\includegraphics[width=1in,height=1.25in,clip,keepaspectratio]{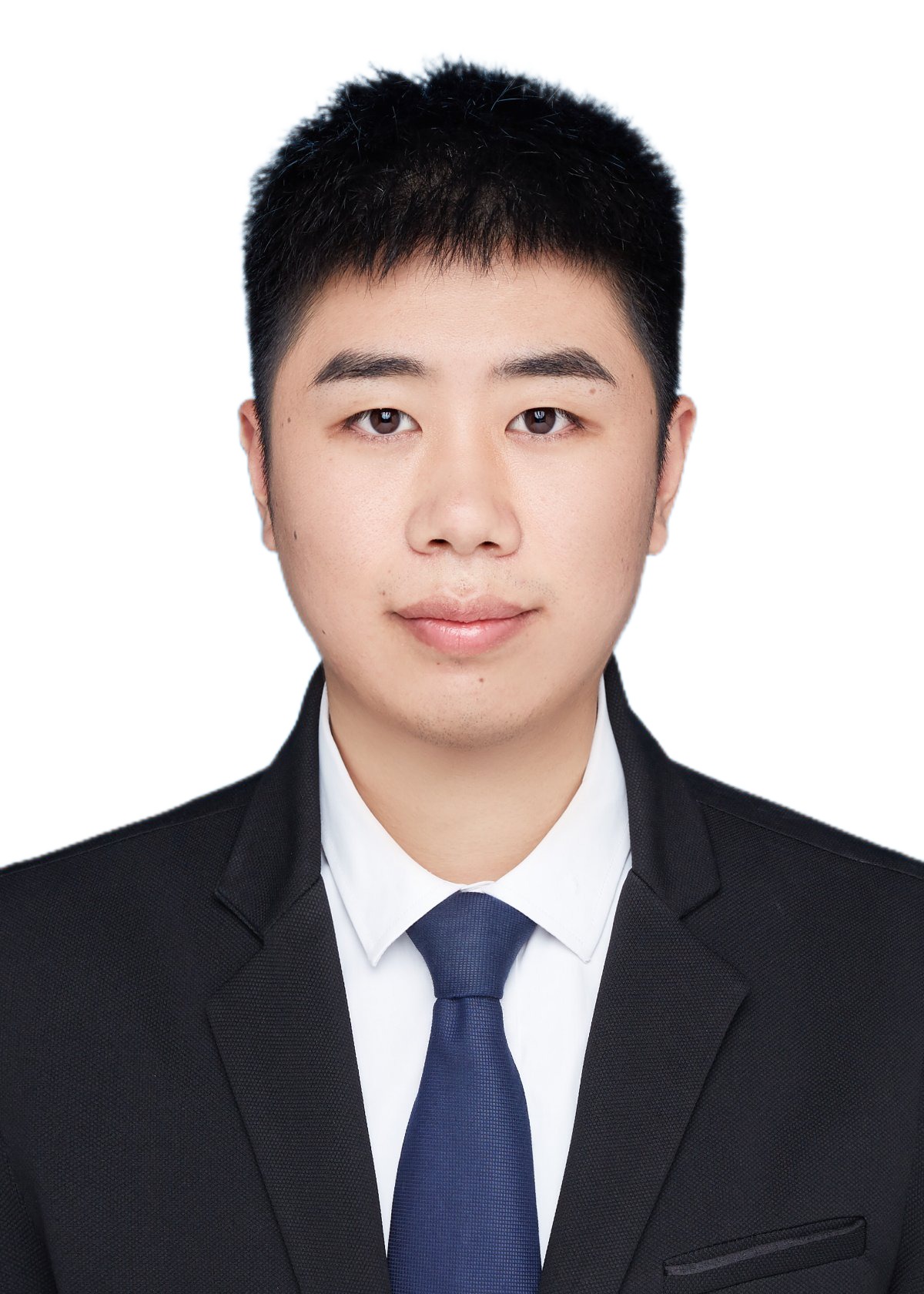}}]{Shuren Qi}
	received the Ph.D. degree in computer science from Nanjing University of Aeronautics and Astronautics, Nanjing, China, in 2024. He is currently a Postdoctoral Fellow with the Department of Mathematics, The Chinese University of Hong Kong, Hong Kong. He has published academic papers in top-tier venues including \emph{ACM Computing Surveys} and \emph{IEEE Transactions on Pattern Analysis and Machine Intelligence}. His research involves the general topics of invariance, robustness, and explainability in computer vision, with a focus on invariant representations, for closing today's trustworthiness gap in artificial intelligence, \emph{e.g.}, forensic and security of visual data.
\end{IEEEbiography}

\begin{IEEEbiography}[{\includegraphics[width=1in,height=1.25in,clip,keepaspectratio]{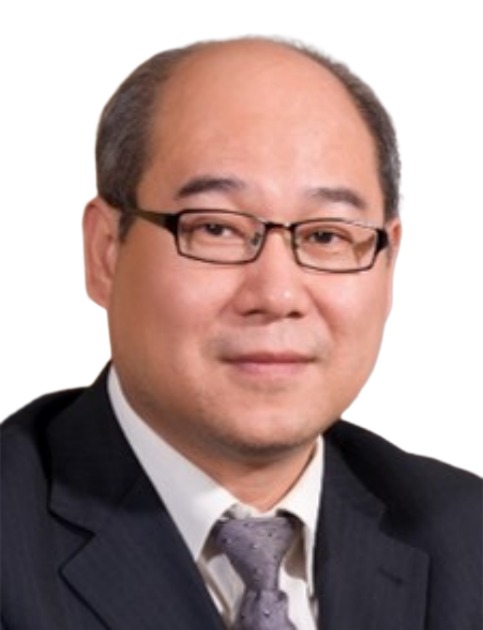}}]{Zhiqiu Huang}
	received the Ph.D. degree in computer science from Nanjing University of Aeronautics and Astronautics, Nanjing, China, in 1999. He is a professor with the College of Computer Science and Technology, Nanjing University of Aeronautics and Astronautics, China. He has authored or coauthored more than 80 journal and conference papers. His research interests include software engineering, formal methods, and knowledge engineering.
\end{IEEEbiography}

\begin{IEEEbiography}[{\includegraphics[width=1in,height=1.25in,clip,keepaspectratio]{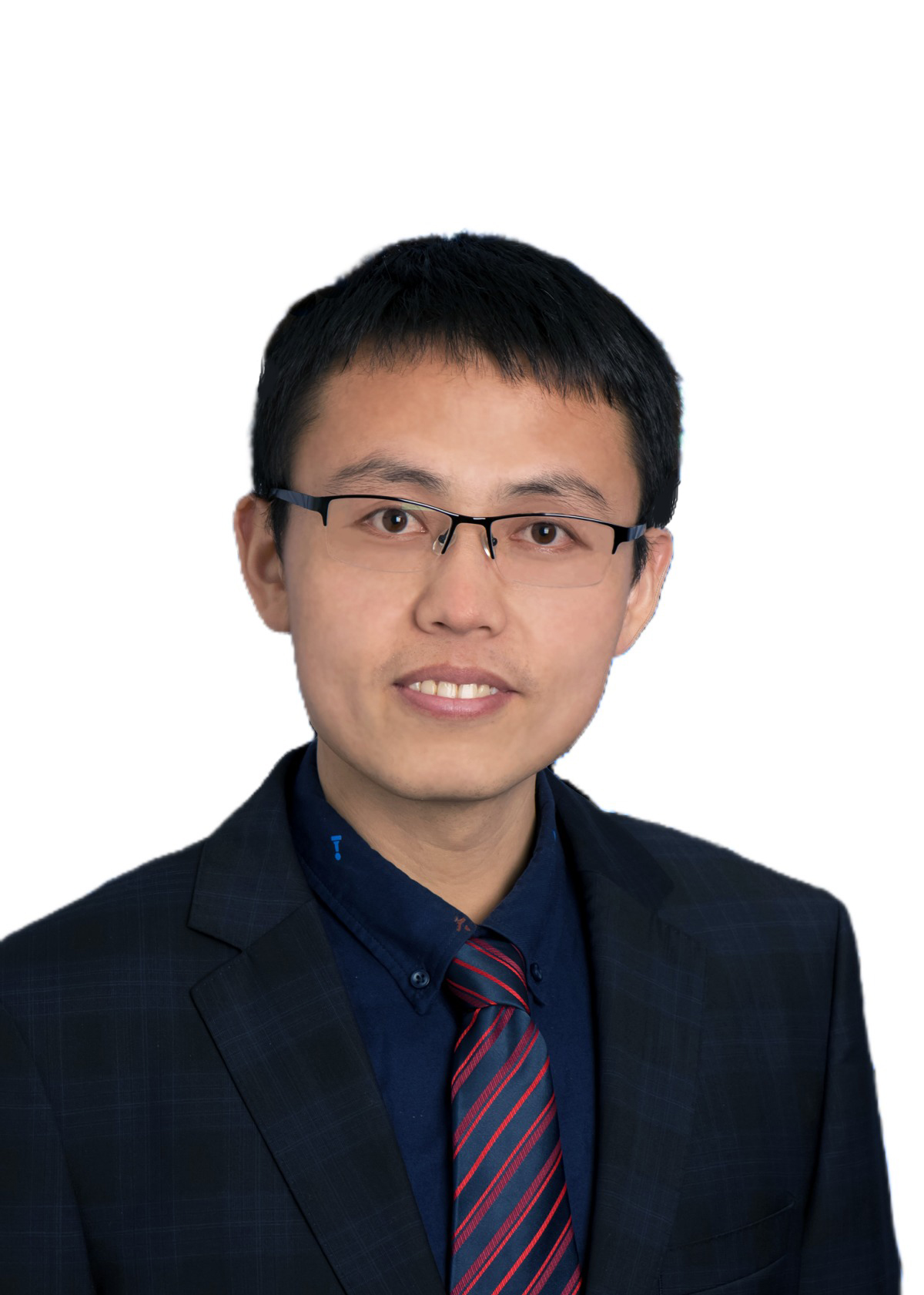}}]{Yushu Zhang}
	(Senior Member, IEEE) received the Ph.D. degree in computer science from Chongqing University, Chongqing, China, in 2014. He held various research positions with the City University of Hong Kong, Southwest University, University of Macau, and Deakin University. He is a Professor with the College of Computer Science and Technology, Nanjing University of Aeronautics and Astronautics, Nanjing, China. His research interests include multimedia processing and security, artificial intelligence, and blockchain. Dr. Zhang is an Associate Editor of \emph{Signal Processing} and \emph{Information Sciences}.
\end{IEEEbiography}

\begin{IEEEbiography}[{\includegraphics[width=1in,height=1.25in,clip,keepaspectratio]{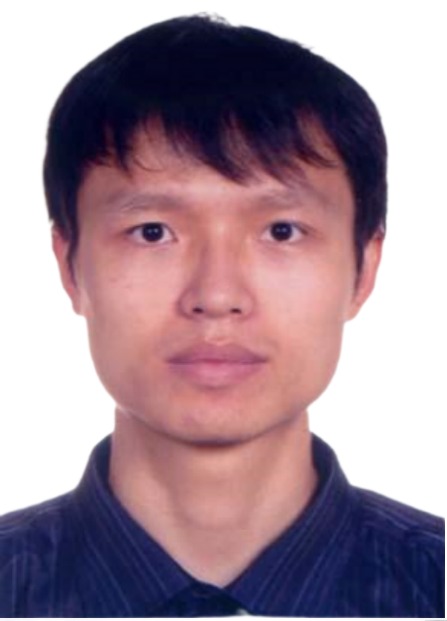}}]{Rushi Lan}
	received the B.S. degree in information and computational science and the M.S. degree in applied mathematics from Nanjing University of Information Science and Technology, Nanjing, China, in 2008 and 2011, respectively, and the Ph.D. degree in soft engineering from University of Macau, Macau, China, in 2016. He is currently a Professor with the Guangxi Key Laboratory of Image and Graphic Intelligent Processing, Guilin University of Electronic Technology, Guilin, China. His research interests include image processing and machine learning.
\end{IEEEbiography}

\begin{IEEEbiography}[{\includegraphics[width=1in,height=1.25in,clip,keepaspectratio]{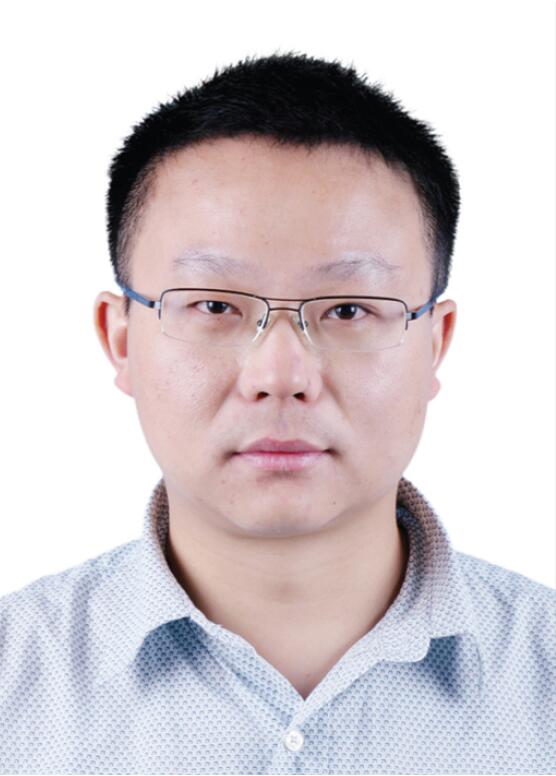}}]{Xiaochun Cao}
	(Senior Member, IEEE) received the B.E. and M.E. degrees in computer science from Beihang University, Beijing, China, in 1999 and 2002, respectively, and the Ph.D. degree in computer science from the University of Central Florida, Orlando, FL, USA, in 2006. After graduation, he spent about three years at ObjectVideo Inc., as a Research Scientist. From 2008 to 2012, he was a Professor at Tianjin University, Tianjin, China. Before joining Sun Yat-sen University, Shenzhen, China, he was a Professor at the Institute of Information Engineering, Chinese Academy of Sciences, Beijing, China. He is a Professor and the Dean with the School of Cyber Science and Technology, Shenzhen Campus of Sun Yat-sen University. He has authored or coauthored more than 200 journal and conference papers. Dr. Cao’s dissertation was nominated for the University Level Outstanding Dissertation Award. He was a recipient of the Piero Zamperoni Best Student Paper Award at the \emph{International Conference on Pattern Recognition}, in 2004 and 2010. He was on the Editorial Boards of \emph{IEEE Transactions on Circuits and Systems for Video Technology} and \emph{IEEE Transactions on Multimedia}. He is on the Editorial Boards of \emph{IEEE Transactions on Pattern Analysis and Machine Intelligence} and \emph{IEEE Transactions on Image Processing}.
\end{IEEEbiography}

\begin{IEEEbiography}[{\includegraphics[width=1in,height=1.25in,clip,keepaspectratio]{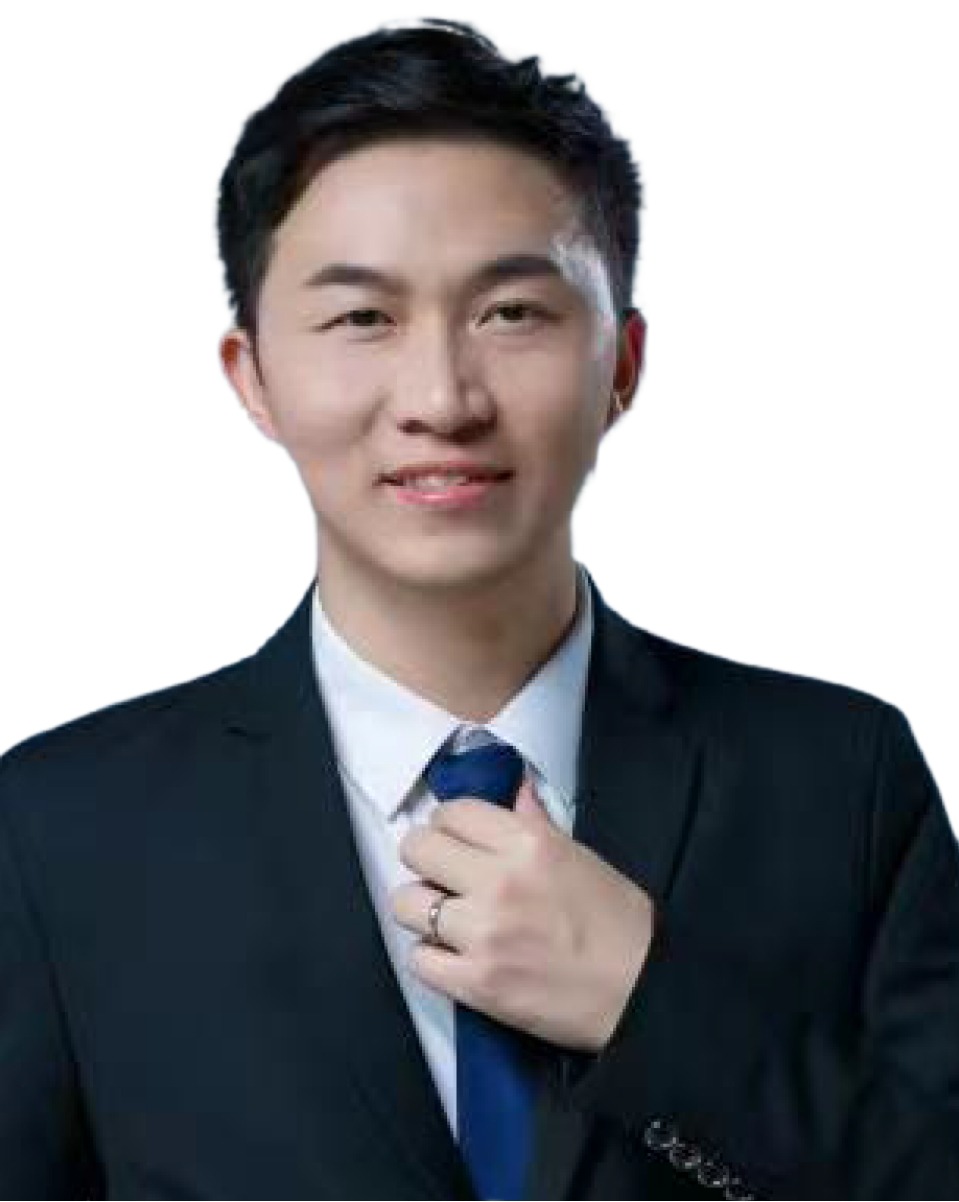}}]{Feng-Lei Fan}
	(Senior Member, IEEE) received the B.S. degree in instrumentation engineering from the Harbin Institute of Technology (HIT), Harbin, China, in 2017, and the Ph.D. degree from Rensselaer Polytechnic Institute (RPI), Troy, NY, USA, in 2021. He is currently a Research Assistant Professor with the Department of Mathematics, The Chinese University of Hong Kong, Hong Kong. He has authored 26 papers in flagship AI and medical imaging journals. His primary research interests lie in deep learning theory and methodology, neuroscience, and medical image processing. Dr. Fan served as a PC Member in many conferences such as the \emph{International Joint Conference on Artificial Intelligence} and the \emph{Association for the Advancement of Artificial Intelligence}. He was a recipient of the 2021 International Neural Network Society Doctoral Dissertation Award.
\end{IEEEbiography}


\vfill


\end{document}